\documentclass{article} % For LaTeX2e
\usepackage[utf8]{inputenc}    % UTF-8 input encoding
\usepackage[T1]{fontenc}       % T1 font encoding for accented characters
\usepackage{nemo_speech,times}

\usepackage{hyperref}       % hyperlinks
\usepackage{url}            % simple URL typesetting
\usepackage{booktabs}       % professional-quality tables
\usepackage{amsfonts}       % blackboard math symbols
\usepackage{nicefrac}       % compact symbols for 1/2, etc.
\usepackage{microtype}      % microtypography
\usepackage{lipsum}
\usepackage{colortbl}       % for \rowcolor
\usepackage{graphicx}       % graphics
\usepackage{multirow}
\usepackage{multicol}
\usepackage{subcaption}
\usepackage{tabularx}
\usepackage{array}
\usepackage{amsmath}
\usepackage{pdflscape}
\usepackage{rotating}
\usepackage{float}
\usepackage{threeparttable} % in preamble
\graphicspath{{images/}}     % organize your images and other figures under media/ folder
\usepackage[table]{xcolor}   % in the preamble

\newcolumntype{Y}{>{\centering\arraybackslash}X}

\title{Canary-1B-v2 \& Parakeet-TDT-0.6B-v3: Efficient and High-Performance Models for Multilingual ASR and AST}

\author{Monica Sekoyan\thanks{These authors contributed equally. Listed in alphabetical order based on First Name.}, Nithin Rao Koluguri\footnotemark[1], Nune Tadevosyan\footnotemark[1], \\
Piotr Zelasko, Travis Bartley, Nikolay Karpov, Jagadeesh Balam, Boris Ginsburg \\
NVIDIA \\
Santa Clara, CA 95051, USA \\
\texttt{\{msekoyan, nkoluguri, ntadevosyan\}@nvidia.com}
}

\nemofinalcopy % Uncomment for camera-ready version, but NOT for submission.

\begin{document}

\maketitle

% Remove the header text completely 
\fancyhead{}

\begin{abstract}
This report introduces \textbf{Canary-1B-v2}\footnote{\url{https://huggingface.co/nvidia/canary-1b-v2}}, a fast, robust multilingual model for Automatic Speech Recognition (ASR) and Speech-to-Text Translation (AST). Built with a FastConformer encoder and Transformer decoder, it supports 25 European languages. The model was trained on ~1.7M hours of total data samples, including Granary and NeMo ASR Set 3.0, with non-speech audio added to reduce hallucinations for ASR and AST. We describe its two-stage pre-training and fine-tuning process with dynamic data balancing, as well as experiments with an nGPT encoder. Results show nGPT scales well with massive data, while FastConformer excels after fine-tuning. For timestamps, Canary-1B-v2 uses the NeMo Forced Aligner (NFA) with an auxiliary CTC model, providing reliable segment-level timestamps for ASR and AST. Evaluations show Canary-1B-v2 outperforms Whisper-large-v3 on English ASR while being 10$\times$ faster, and delivers competitive multilingual ASR and AST performance against larger models like Seamless-M4T-v2-large and LLM-based systems.

We also release \textbf{Parakeet-TDT-0.6B-v3}\footnote{\url{https://huggingface.co/nvidia/parakeet-tdt-0.6b-v3}}, a successor to v2, offering multilingual ASR across the same 25 languages with just 600M parameters.

\end{abstract}

% keywords can be removed
%\keywords{First keyword \and Second keyword \and More}

\section{Introduction}

% \section{Introduction}

Modern speech processing systems for Automatic Speech Recognition (ASR) and Speech Translation (ST) are predominantly built on the encoder-decoder paradigm. Architectures such as the Transformer\cite{attention} have proven highly effective, with models like Whisper\cite{radford2023whisper} and SeamlessM4T\cite{barrault2023seamlessm4t} demonstrating that training on massive, weakly supervised multilingual datasets can yield state-of-the-art results. Despite their strong performance, these models often require large resources and run slowly, creating a trade-off between accuracy, size, and speed. Specialized architectures like Conformer and its efficient variant, FastConformer, have been developed to better capture speech-specific features with a smaller computation footprint.

This work introduces Canary-1B-v2, a multilingual, multi-task model designed to deliver robust ASR and AST performance with high efficiency. In addition, we release Parakeet-TDT-0.6B-v3, a smaller yet more accurate ASR model supporting the same 25 languages.

The key contributions of this paper are:

\begin{itemize}
\item \textbf{A Multi-Stage Training and Fine-Tuning Strategy}: A two-stage pre-training regimen followed by a high-quality fine-tuning stage with dynamic weight scheduling to address significant data imbalances in large-scale multilingual corpora.

\item \textbf{Comprehensive Data Curation}: Training on a 1.7 million-hour dataset spanning 25 languages, combining pseudo-labeled and human-annotated data, and incorporating non-speech audio to improve robustness.

\item \textbf{Efficient Architecture and Timestamp Generation}: Canary-1B-v2 leverages a FastConformer encoder and a unified BPE tokenizer for all languages, delivering high throughput. It integrates the NeMo Forced Aligner (NFA) with an auxiliary CTC model for accurate segment-level timestamps in both ASR and AST tasks.

\item \textbf{State-of-the-Art Performance}: Canary-1B-v2 achieves competitive or superior results to much larger models on ASR and AST benchmarks, including the Hugging Face Open ASR Leaderboard, while providing significantly faster inference.

\end{itemize}

This report details the models’ architectures, the data curation and balancing strategies, the multi-stage training process, and extensive evaluations against leading speech processing systems.

% \textbf{Need to combine models and how good they are with prior work}

\section{Architecture}
% \section{Architecture}
\label{sec:arch}
Modern speech recognition and speech translation systems are predominantly built on the encoder-decoder paradigm. In this framework, the encoder extracts high-level representations from the raw speech signal, conditioning the decoder to generate text in the target language. The most powerful recent systems employ pure Transformer-based architectures, often trained at a massive speech scale data. For instance, Whisper \cite{radford2023whisper} demonstrated that training a Transformer encoder-decoder on hundreds of thousands of hours of weakly supervised multilingual data yields state-of-the-art performance on both Automatic Speech Recognition (ASR) and Speech Translation (ST) tasks. Similarly, SeamlessM4T \cite{barrault2023seamlessm4t} extended this approach to create a unified multilingual framework for speech and text translation, underscoring the scalability of Transformer-based designs. Earlier models, such as the Speech-Transformer \cite{dong2018speech}, were instrumental in establishing the viability of attention-based encoder-decoder architectures for speech processing.

In contrast to general-purpose Transformers, several architectures have been tailored specifically for speech. The Conformer \cite{gulati2020conformer} augments the Transformer with convolutional modules, enabling it to capture both fine-grained local acoustic patterns and long-range global dependencies. Its optimized variant, the FastConformer \cite{rekesh2023fastconformer}, further enhances efficiency through aggressive subsampling and the use of depthwise separable convolutions. Such encoders are often paired with online decoders like the RNN-Transducer (RNN-T) \cite{jain2019rnntransducer}, Token and Duration Transducer (TDT)\cite{xu2023efficient} or Connectionist Temporal Classification (CTC) \cite{graves2006connectionist}, which are widely adopted for streaming ASR due to their low-latency characteristics.

For speech translation, however, autoregressive Transformer-based decoders remain dominant, as they excel at modeling cross-lingual dependencies and generating fluent text. Recently, a prominent strategy has been to integrate pre-trained Large Language Models (LLMs) as decoders. This approach leverages the vast linguistic knowledge captured during text-only pre-training and scales effectively across modalities. For example, models like SALM\cite{chen2024salm}\footnote{\url{https://huggingface.co/nvidia/canary-qwen-2.5b}}, BESTOW \cite{chen2024bestow} and SpeechGPT \cite{zhang2023speechgpt} adapt speech encoders to interface with powerful LLMs. One example of such integration is Phi-4-Multimodal \cite{abouelenin2025phi4multimodal}. In this architecture, inputs from multiple modalities—including speech—are processed by modality-specific encoders (e.g., a Conformer for speech). The resulting features are projected into a common embedding space via lightweight adapters and then decoded by a shared LLM backbone (Phi-4-Mini) equipped with LoRA adapters for speech-to-text tasks. This design circumvents traditional decoder retraining by adapting the LLM with LoRA, efficiently enabling language generation conditioned on speech inputs.

 Notable approach from above models is the Speech-Augmented Language Model (SALM) \cite{chen2023salm}, which utilizes a frozen pre-trained text LLM and augments it with a speech encoder, a modality adapter, and LoRA layers. This creates a unified architecture for both ASR and ST. Trained with in-context speech instruction tuning, SALM matches the performance of specialized Conformer models while also demonstrating zero-shot adaptation and keyword boosting capabilities. Such designs not only improve translation quality but also align speech models more closely with the architectural principles of today's leading text-based systems.

\subsection{Encoder}
In any encoder-decoder architecture for speech, the encoder is a critical component. It is typically responsible for the majority of the model's memory consumption. The full self-attention mechanism in encoder layers often becomes the primary bottleneck during pre-training and inference, especially with long audio sequences. Consequently, the encoder's design plays a central role in determining both the scalability and efficiency of the entire system. Even minor architectural modifications or scaling decisions can yield substantial performance gains. In this work, we focus on the encoder and experiment with two distinct architectures: the FastConformer \cite{rekesh2023fast} and the normalized GPT (nGPT) \cite{loshchilov2024ngpt}.

\subsubsection{FastConformer}
The FastConformer is an evolution of the Conformer encoder, explicitly optimized for computational efficiency in speech tasks. Its primary improvements over the original Conformer include:

\begin{itemize}
    \item \textbf{Aggressive Subsampling (8x):} Input features are downsampled by a factor of 8 at the input stage using convolutional blocks. This significantly shortens the sequence length, reducing computational cost while preserving accuracy.
    \item \textbf{Depthwise Separable Convolutions:} Standard convolutions in both the subsampling layers and the Conformer blocks are replaced with depthwise separable convolutions, which reduces the number of operations while maintaining strong performance.
    \item \textbf{Lightweight Convolutional Modules:} The convolutional kernel size is reduced (e.g., from 31 to 9) and channel dimensions are scaled down, further streamlining the encoder without degrading recognition quality.
\end{itemize}

These optimizations yield a model that is significantly faster (2–3× inference speedup) and more memory-efficient, while achieving accuracy comparable to or better than the original Conformer.

\vspace{0.3em} % space before figure
\begin{figure}[htbp]
    \centering
    \includegraphics[width=0.8\textwidth]{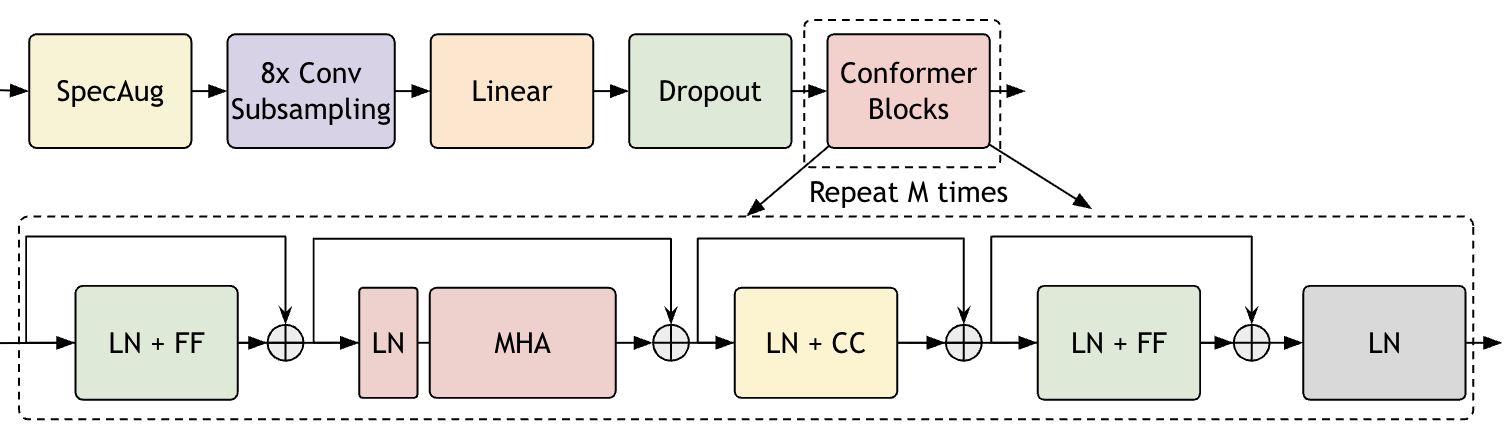}
    \caption{Overview of the \textbf{FastConformer} architecture. The input features are first subsampled to a resolution of 80\,ms, followed by a linear projection and dropout for dimensionality reduction and regularization. The core of the model consists of a stack of repeated Conformer blocks, each containing four main components: Layer Normalization (LN), a Feed-Forward (FF) module, a Multi-Head Attention (MHA) mechanism, and a Convolutional Module (CC).}
    \label{fig:fastconformer}
\end{figure}

\subsubsection{nGPT}
While the FastConformer optimizes a speech-specific architecture, a different line of research has focused on fundamentally rethinking the Transformer itself. Normalized GPT (nGPT) model \cite{loshchilov2024ngpt}  introduces a hyperspherical normalization scheme that constrains all embeddings to a unit hypersphere. This design stabilizes training, improves generalization, and accelerates convergence, enabling the model to reach target validation metrics in fewer steps than standard Transformers.

Building on these ideas, we adapt nGPT into a full encoder-decoder architecture for speech. To the best of our knowledge, this is the first work to explore the nGPT architecture for speech processing tasks. As an initial step, we developed and evaluated an nGPT-based encoder, allowing us to isolate and study its effectiveness in the acoustic domain. For the decoder, we retained a conventional Transformer architecture to ensure that any observed improvements could be directly attributed to the novel encoder design.

Our implementation of the \texttt{NGPTEncoder} begins with a Linear subsampling front-end to reduce the temporal resolution of the input spectrogram. The encoder is then constructed as a stack of nGPT layers, with the following key components:

\begin{itemize}
    \item \textbf{Multi-head Attention with Rotary Embeddings.} The attention layers use Rotary Positional Embeddings (RoPE) for long-sequence generalization and standard scaled dot-product attention.
    \item \textbf{Feed-forward Network.} Each layer uses a gated two-stage feed-forward module with a SiLU activation.
    \item \textbf{Hyperspherical Normalization.} All embeddings and intermediate activations are constrained to lie on a unit hypersphere. Query, key, and projection weights are explicitly normalized, with learnable scaling factors ($\alpha, \sigma$) regulating the residual contributions from the attention and feed-forward modules.
    \item \textbf{Residual Connections.} Residual paths can be interpreted as a first-order optimization step, where each layer refines the representation by adding an incremental update to the current state.
\end{itemize}

\vspace{0.3em} % space before figure
\begin{figure}[htbp]
    \centering
    \includegraphics[width=0.5\textwidth]{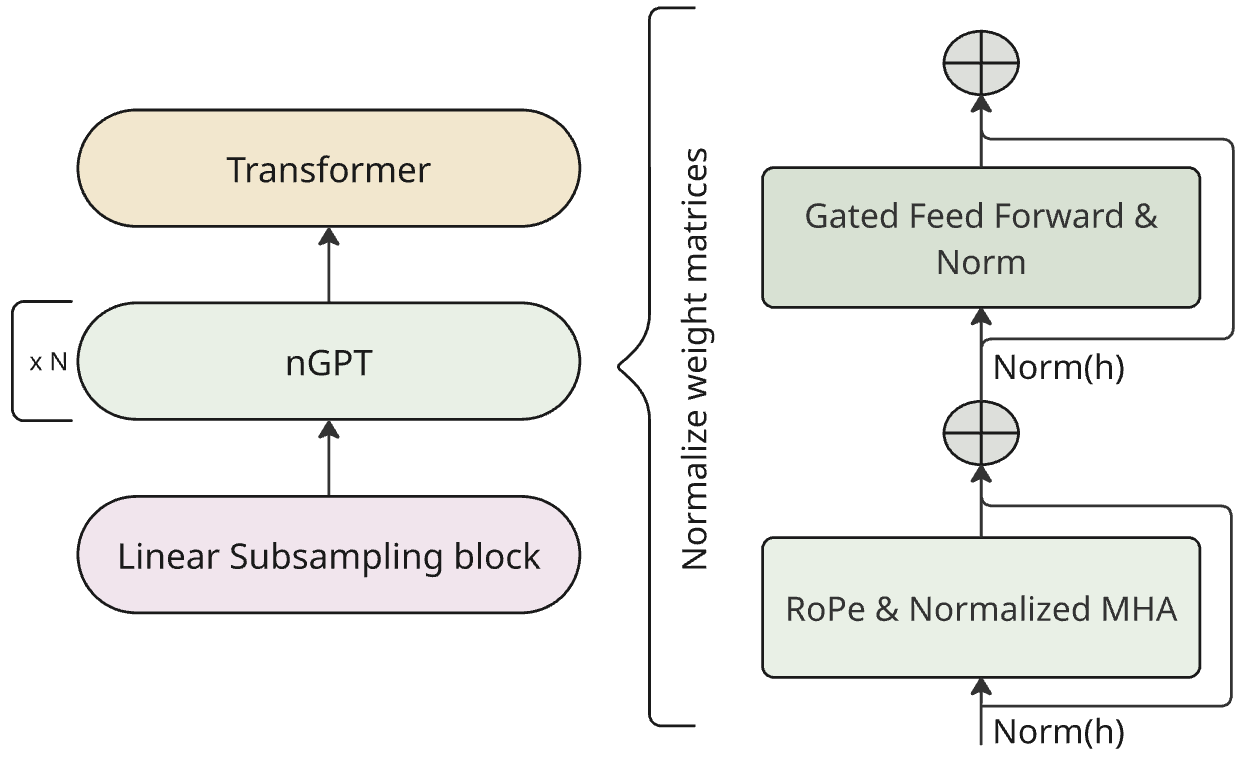}
    \caption{Overview of the nGPT architecture. The input features are first processed by a linear subsampling block. This is followed by a stack of repeated nGPT layers and a Transformer decoder. Each nGPT layer adopts a Transformer-style design with Rotary Position Embeddings (RoPE), multi-head attention, and gated feed-forward modules. Normalization is applied across the embedding dimension, and both activations and weight matrices are normalized, with weight normalization additionally enforced after each optimizer update.}
    \label{fig:ngpt_architecture}
\end{figure}

\subsubsection{Positional Encodings}
Our initial nGPT implementation adopted Rotary Positional Encodings (RoPE) \cite{su2021roformer} as the baseline. RoPE injects position information by applying a rotation to the query and key vectors, a design known to generalize well in long-context settings. While functional, our experiments with RoPE revealed performance degradation in very long-context ASR and ST scenarios.

To address this limitation, we explored Attention with Linear Biases (ALiBi) \cite{press2022alibi}. ALiBi was originally designed for autoregressive language models, where it adds a static, position-dependent bias to the attention scores, penalizing attention to distant tokens. Since our encoder processes the full context (i.e., it is bidirectional), we modified the original ALiBi formulation to use a symmetric bias matrix, ensuring that both left and right contexts are treated equally. This adaptation allowed us to integrate ALiBi into our nGPT encoder as an alternative positional encoding scheme. This represents the first exploration of ALiBi positional embeddings for speech recognition and translation tasks.

\vspace{0.3em} % space before figure
\begin{figure}[htbp]
    \centering
    \includegraphics[width=0.2\textwidth]{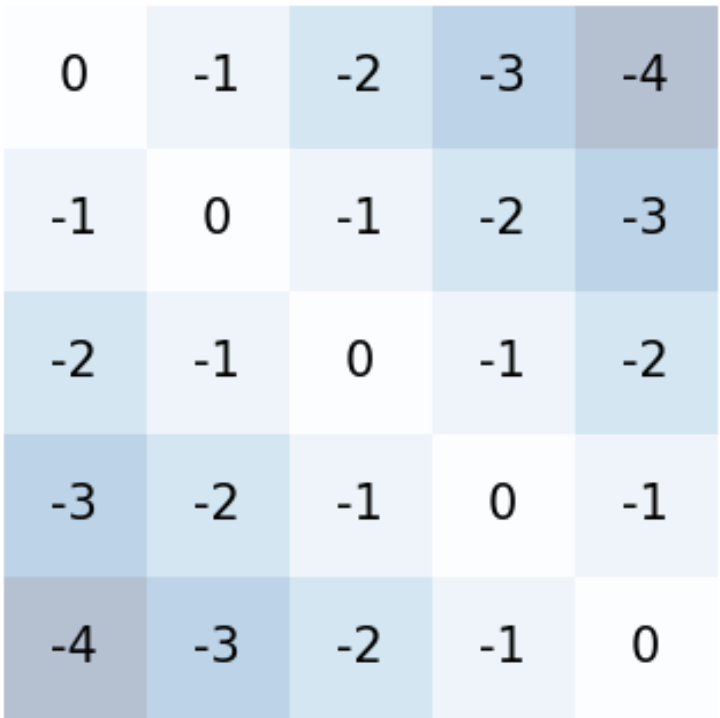}
    \caption{Symmetric ALiBi bias matrix used in the nGPT encoder. Unlike the original causal ALiBi, which penalizes only future positions, this non-causal variant applies equal bias to tokens before and after the current position, ensuring balanced treatment of left and right context.}

    \label{fig:alibi_matrix}
\end{figure}

\section{Data}
% \section{Data}
\label{sec:data}

Web-scale training data has proven transformative for multi-task and multi-lingual speech systems, delivering substantial performance improvements. However, the scale that enables these gains often comes with quality trade-offs that require careful management. For Canary-1B-v2 and Parakeet-TDT-0.6B-v3, we addressed the data problem from multiple complementary perspectives: task and language balancing, optimal mixing of pseudo-labeled and human-annotated data, and careful curation strategies to maximize data utility while mitigating quality concerns. 

In this section, we present all the datasets that comprise our 1.7M hours of training data used to train both models, provide insights into data composition and future balancing decisions, and detail the evaluation datasets used to assess model performance.

\subsection{Training Data Curation and Sources }
\label{sec:data-training-data}

The training data of \emph{Canary-1B-v2} is heavily based on Granary multilingual and multitask dataset \cite{granary} which is a super set of three main corpora: MOSEL\cite{mosel}, YTC\cite{ytc}, and YODAS\cite{yodas}, comprising around ~0.7 million hours of ASR speech data samples in 25 supported languages, which is pseudo-labeled and translated to other languages for AST tasks with Granary pipeline. Its multi-task speech data components, proven quality of the data processing pipeline, and broad language coverage made it possible to train a model of this scale and scope. However, because much of the data was pseudo-labeled, we also needed to mitigate potential issues that could arise from the limited amount of human-labeled data.

To address this challenge, we augmented Granary with 227,000 hours of high-quality human-labeled data from NeMo ASR Set 3.0, covering all three tasks (ASR, AST X→En, and En→X) across all 25 languages. While representing only 13\% of our total training dataset, NeMo ASR Set provides invaluable human-annotated data sourced from proven quality datasets including AMI~\cite{ami}, FLEURS~\cite{fleurs}, Common Voice~\cite{commonvoice:2020}, Multilingual LibriSpeech (MLS)~\cite{MLS}, and language-specific datasets such as Golos~\cite{golos} for Russian.

Another crucial addition to the Granary dataset was a supplementary dataset specifically for the En→X translation task, since Granary originally provided data only for ASR and X→En tasks. This supplementary mono-task dataset was constructed by sampling a subset of the Granary English corpus and generating translations into multiple target languages using the methodology described in \cite{granary}. In total, it comprised around 480,000 hours of English-to-non-English audio–transcription pairs.

\begin{figure}[htbp]
  \centering
  \includegraphics[width=0.6\linewidth]{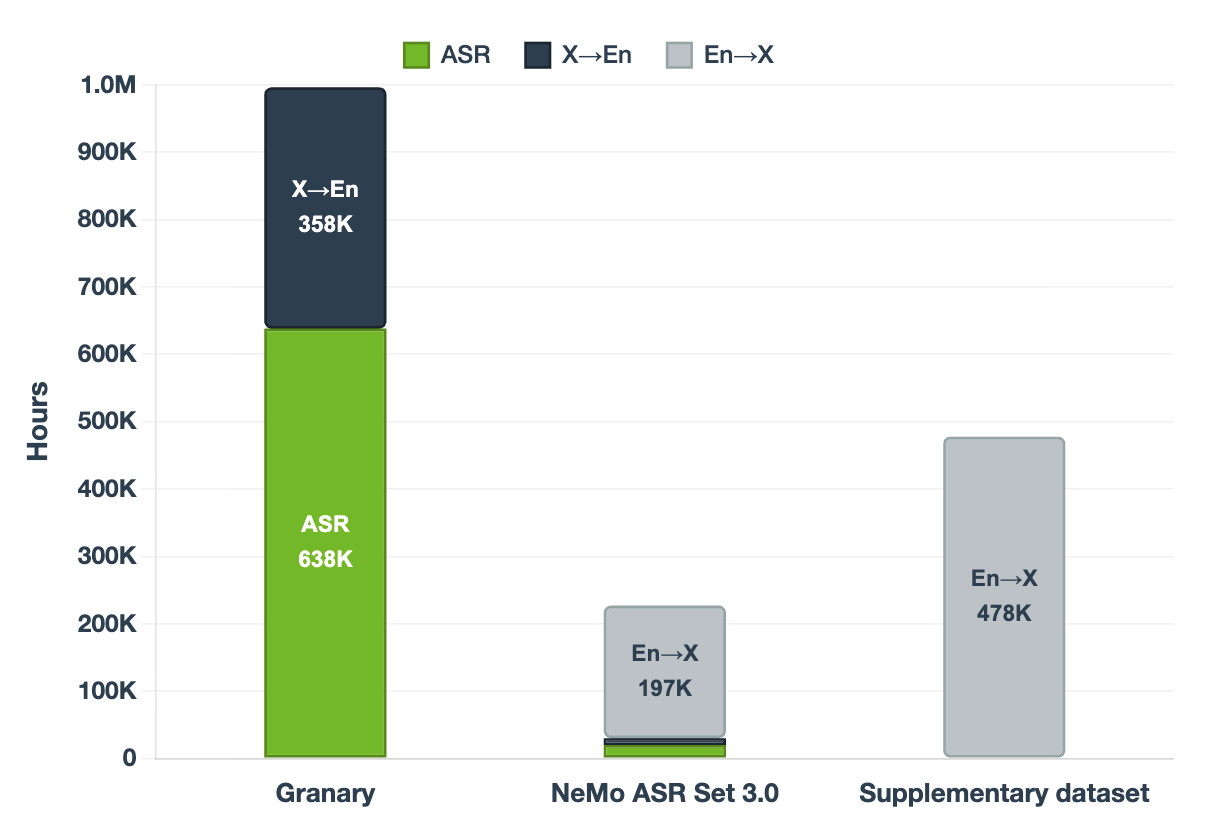}
  \caption{Overview of participating corpora in the training set with within-corpus task divisions.}
  \label{fig:data-dist}
\end{figure}

Figure~\ref{fig:data-dist} summarizes the distribution of tasks across corpora in our training data. As can be observed, there is a task imbalance present in the training set, where ASR and En→X tasks dominate the dataset, while X→En translation data is comparatively limited. This is a common phenomenon in multi-task training setups, for which we have applied mitigating methods described in Section~\ref{sec:training}.

For our second model, \emph{Parakeet-TDT-0.6B-v3}, we trained exclusively on the ASR subset of the \emph{Canary-1B-v2} training set (\(\sim\)660,000 hours across the same 25 languages).

\subsection{Non-speech Data for Robustness}

To enhance model robustness against hallucinations, particularly those potentially inherited from Whisper-based pseudo-labeling, we incorporated 36,000 hours of non-speech audio into our training dataset. These samples, extracted during Granary pipeline filtration steps\footnote{\url{https://github.com/NVIDIA/NeMo-speech-data-processor/tree/main/dataset_configs/multilingual/granary}}, are paired with empty string targets to teach the model when not to generate transcriptions.

We randomly assigned source-target language pairs to non-speech samples, creating coverage across all possible task combinations (X→X, En→X, and X→En). This approach ensures the model learns to detect and appropriately handle non-speech content regardless of the specified language pair.

With this final addition, we have around 1.7M hours of combined ASR and AST data. For a more detailed overview of the training data decomposition, refer to Appendix~\ref{app:training_data_overview}.

\subsection{Language Coverage and Imbalance Analysis}
\label{sec:lang-coverage-imbalance}

Our training dataset encompasses all 25 target European languages, with Granary serving as the primary source for low-resource languages such as Latvian (lv), Maltese (mt), and Lithuanian (lt). Through VoxPopuli~\cite{voxpopuli} from MOSEL, we successfully gathered sufficient training material for these languages and generated corresponding X→En translation pairs leveraging Granary pipeline.

English naturally dominates the dataset, comprising approximately 285,000 hours (~40\%) of ASR data. Given English's central role across all three tasks (ASR, X→En, En→X), this dominance supports building a solid foundational knowledge base during pretraining. 

\begin{figure}[htbp]
  \centering
  \includegraphics[width=0.95\linewidth]{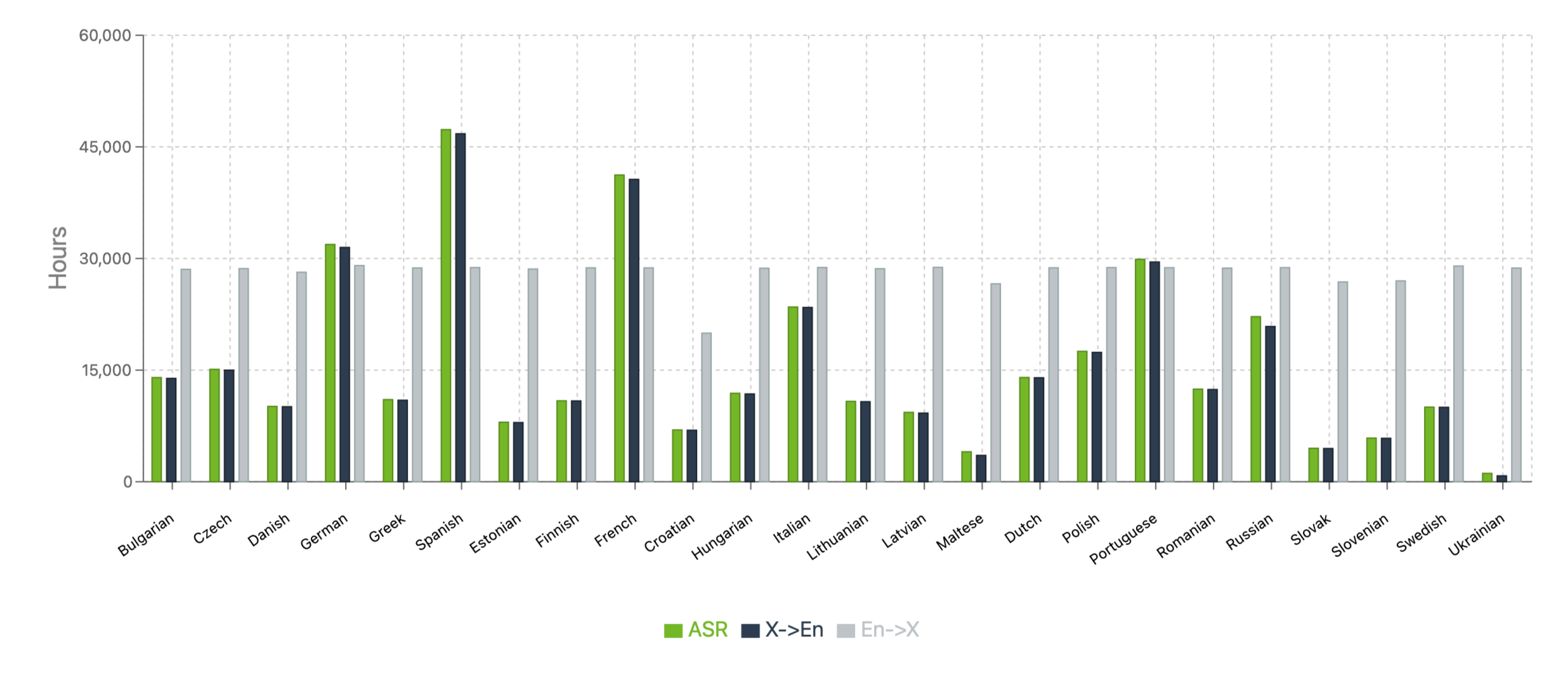}
  \caption{Language (non-English) Training Hours Distribution by Task.}
  \label{fig:language-distribution}
\end{figure}

Figure~\ref{fig:language-distribution} presents the training hours distribution for our 24 non-English target languages, revealing important patterns in data availability across tasks. Language imbalance mitigation strategies, including approaches to address English dominance, are detailed in Sections ~\ref{sec:lang-corpora-balancing} and ~\ref{sec:fine-tuning}.

\begin{figure}[htbp]
  \centering
  \includegraphics[width=0.95\linewidth]{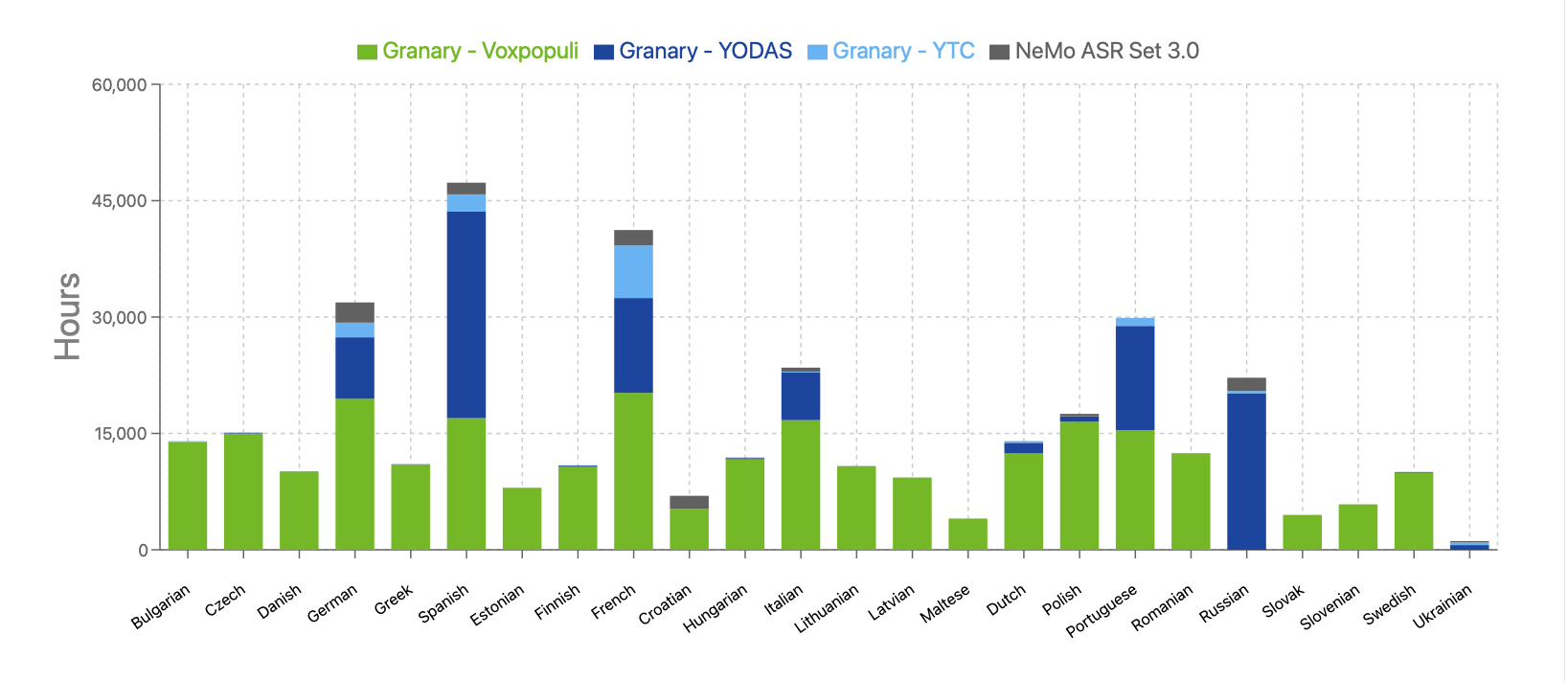}
  \caption{Language (non-English) ASR Data Training Hours Distribution by Corpus.}
  \label{fig:language-corpus}
\end{figure}

As observed in the Figure~\ref{fig:language-distribution}, the En→X translation direction exhibits relatively uniform data distribution across languages, with values consistently ranging between 20,000-29,000 hours. The primary divergence emerges in ASR and X→En translation tasks, where low-resource languages such as Ukrainian (790 hours), Maltese, Estonian, and Slovak contrast sharply with high-resource languages including Spanish, French, German, Italian, Portuguese, and Russian, which dominate with significantly higher training hours. Given this pronounced imbalance and the fact that VoxPopuli data from the Granary corpus serves as the primary ASR source, we further analyze the ASR data source composition in the subsequent Figure~\ref{fig:language-corpus}.

When examining the relative volumes of Granary sub-corpora and NeMo ASR Set 3.0 in the ASR data for each language under study, a clear trend emerges. Except for high-resource, well-represented languages such as German, Spanish, French, Italian, Portuguese, and Russian, the Voxpopuli subset of the Granary corpus dominates, comprising over 95\% of the data for most languages. This dominance of Voxpopuli creates two primary issues for these languages: (a) Voxpopuli recordings originate from EU parliament sessions, where the language register differs significantly from everyday conversational speech, leading to narrow-domain data dominance; and (b) Voxpopuli consists primarily of 30-second audio segments, which limits the acoustic and linguistic diversity available for languages dominated by this corpus. We reserve Voxpopuli audio resegmentation for future work. Despite these limitations, we cannot eliminate Voxpopuli entirely, as doing so would remove 15 unique languages from our 25-language set, contradicting our objective of supporting low-resource language inclusion.

For detailed numerical breakdowns of training hours by language, corpus, and task, refer to Appendix~\ref{app:lang_dist}.

\subsubsection{Balancing Languages and Corpora}
\label{sec:lang-corpora-balancing}

Building on the preceding analysis of language and corpora imbalance, we adopt a two-tier sampling policy that (i) balances corpora \emph{within} each language and (ii) balances \emph{across} languages. During pre-training, each direction is treated as a distinct “language” entry (e.g., \texttt{lv}, \texttt{lv--en}, \texttt{en--lv}), so the policy applies uniformly to ASR and AST.  

We followed the general methodology described in \cite{xlsr}, with one key modification. In \cite{xlsr}, the procedure first upsamples languages within a corpus and then balances the different corpora by treating each corpus as a separate language. In contrast, our approach inverts this order: we first perform corpus balancing within each language, and only then apply language balancing across the training corpora. This adjustment ensures that low-resource corpora within a language are not overshadowed before cross-language balancing is applied. Our procedure consists of three steps:

\begin{enumerate}
  \item \textbf{Corpus balancing within language.}  
  For a given language $l$, we first compute unnormalized weights for each corpus $c \in l$:
  \[
  w_c = \left(\frac{n(c)}{N_l}\right)^{\alpha},
  \]
  where $n(c)$ is the number of hours in corpus $c$, $N_l = \sum_{c \in l} n(c)$ is the total number of training hours for language $l$, and $\alpha \in (0,1]$ is a subsampling factor controlling the trade-off between high-resource and low-resource corpora.  
  The normalized distribution over corpora within language $l$ is then
  \[
  p_c = \frac{w_c}{\sum_{c \in l} w_c}, \quad \sum_{c \in l} p_c = 1 .
  \]

  \item \textbf{Language balancing across training corpora.}  
  Next, we compute unnormalized weights for languages:
  \[
  w_l = \left(\frac{n(l)}{N_\text{total}}\right)^{\beta},
  \]
  where $n(l) = \sum_{c \in l} n(c)$ is the total number of hours for language $l$, $N_\text{total} = \sum_l n(l)$ is the global total across all languages, and $\beta \in (0,1]$ is an upsampling factor controlling the degree of balance between high- and low-resource languages.  
  The normalized distribution over languages is then
  \[
  p_l = \frac{w_l}{\sum_{l'} w_{l'}}, \quad \sum_l p_l = 1 .
  \]

  \item \textbf{Final entry distribution.}  
  The final probability of sampling from a corpus $c$ in language $l$ is the product
  \[
  p_{c,l} = p_l \cdot p_c .
  \]
\end{enumerate}

In practice, we set both the corpus-level subsampling factor and the language-level upsampling factor to $\alpha = 0.5$ and $\beta = 0.5$, respectively. Having computed these distributions, we leveraged the Lhotse multi-functional library \cite{lhotse} to form batches according to $p_{c,l}$. This ensured diverse sampling at each step. Our analysis showed that, despite the inherent imbalance across languages and tasks, the upsampling procedure consistently produced batches containing at least 14 distinct language pairs.

\subsection{Evaluation Benchmarks}
\label{sec:evaluation-benchmarks}

Having established our training data composition, we now describe the evaluation benchmarks used to assess model performance across tasks and languages.

As canary-1b-v2 is multi-task model, it required comprehensive multi-task evaluation sets for assessment. For this purpose, we primarily used the FLEURS dataset, which included all 25 languages of our focus. In addition to the FLEURS dataset, we also utilized CoVoST\cite{covost2} and Multilingual LibriSpeech (MLS) data. The latter two datasets did not represent all 25 languages, but rather covered a subset of them. See Table~\ref{tab:evaluation-datasets} for complete details on the evaluation datasets and their language coverage.

\begin{table}[ht]
\centering
\small
\setlength{\tabcolsep}{6pt}
\setlength{\abovecaptionskip}{10pt} % add space between table and caption
\renewcommand{\arraystretch}{1.15}
\begin{tabular}{l|ccc|cc|cc}
\toprule
 & \multicolumn{3}{c|}{\textbf{ASR}} & \multicolumn{2}{c|}{\textbf{X$\to$En}} & \multicolumn{2}{c}{\textbf{En$\to$X}} \\
\textbf{Lang} & Fleurs & MLS & CoVoST2 & Fleurs & CoVoST2 & Fleurs & CoVoST2 \\
\midrule
bg & \checkmark & -- & -- & \checkmark & -- & \checkmark & -- \\
cs & \checkmark & -- & -- & \checkmark & -- & \checkmark & -- \\
da & \checkmark & -- & -- & \checkmark & -- & \checkmark & -- \\
de & \checkmark & -- & \checkmark & \checkmark & \checkmark & \checkmark & \checkmark \\
el & \checkmark & -- & -- & \checkmark & -- & \checkmark & -- \\
en & \checkmark & -- & \checkmark & -- & -- & -- & -- \\
es & \checkmark & \checkmark & \checkmark & \checkmark & \checkmark & \checkmark & -- \\
et & \checkmark & -- & \checkmark & \checkmark & \checkmark & \checkmark & \checkmark \\
fi & \checkmark & -- & -- & \checkmark & -- & \checkmark & -- \\
fr & \checkmark & \checkmark & \checkmark & \checkmark & \checkmark & \checkmark & -- \\
hr & \checkmark & -- & -- & \checkmark & -- & \checkmark & -- \\
hu & \checkmark & -- & -- & \checkmark & -- & \checkmark & -- \\
it & \checkmark & \checkmark & \checkmark & \checkmark & \checkmark & \checkmark & -- \\
lt & \checkmark & -- & -- & \checkmark & -- & \checkmark & -- \\
lv & \checkmark & -- & \checkmark & \checkmark & \checkmark & \checkmark & \checkmark \\
mt & \checkmark & -- & -- & \checkmark & -- & \checkmark & -- \\
nl & \checkmark & \checkmark & \checkmark & \checkmark & \checkmark & \checkmark & -- \\
pl & \checkmark & \checkmark & -- & \checkmark & -- & \checkmark & -- \\
pt & \checkmark & \checkmark & \checkmark & \checkmark & \checkmark & \checkmark & -- \\
ro & \checkmark & -- & -- & \checkmark & -- & \checkmark & -- \\
ru & \checkmark & -- & \checkmark & \checkmark & \checkmark & \checkmark & -- \\
sk & \checkmark & -- & -- & \checkmark & -- & \checkmark & -- \\
sl & \checkmark & -- & \checkmark & \checkmark & \checkmark & \checkmark & \checkmark \\
sv & \checkmark & -- & \checkmark & \checkmark & \checkmark & \checkmark & \checkmark \\
uk & \checkmark & -- & \checkmark & \checkmark & -- & \checkmark & -- \\
\bottomrule
\end{tabular}
\caption{Language coverage by task and dataset in evaluation set. A checkmark indicates that language is present in the corresponding dataset for the given task.}
\label{tab:evaluation-datasets}
\end{table}

In addition to these multilingual benchmarks, we evaluated our model on the Hugging Face Open ASR Leaderboard datasets\cite{open-asr-leaderboard} for comprehensive English ASR assessment and competitive benchmarking against other state-of-the-art models.

\section{Training}
% \section{Training}
\label{sec:training}

The development of state-of-the-art multi-task and multi-language models necessitates a training paradigm that goes beyond traditional methods. Modern large-scale training is typically a two-stage process that separates the acquisition of broad, general knowledge from the refinement of task-specific skills.

The first phase, pre-training, is foundational: the model is exposed to vast amounts of unlabeled or weakly supervised data. The goal is to capture fundamental statistical patterns of language and speech, thereby establishing a robust general foundation. Without this stage, training each new task from scratch would demand prohibitively large labeled datasets and computational resources.

The second phase, fine-tuning, adapts this general foundation to specific tasks or domains using smaller, task-oriented labeled datasets. This stage allows for targeted improvements, making the model more effective in practical scenarios. In our work, we employ this two-stage paradigm as well, leveraging fine-tuning as a mitigation strategy for data-related challenges.

A crucial component underpinning both stages is the tokenizer, which determines how raw text or speech transcriptions are decomposed into units that the model can process. The quality and design of the tokenizer directly affect training efficiency and downstream performance, especially in multilingual and multi-task contexts \cite{abagyan2025tokenizerruleallemergent}. Because of this central role, tokenizer training merits dedicated discussion, which we provide in the next subsection.

\subsection{Tokenizer Training}

Unlike the initial versions of the Canary model, which employed concatenated tokenizers \cite{concatenated_tokenizer}, we trained a unified tokenizer for all 25 included languages in this work. This decision was motivated by three key factors. First, preliminary experiments comparing unified and concatenated tokenizers in identical setups demonstrated that the unified tokenizer significantly outperformed the concatenated approach. Second, a unified tokenizer enables natural code-switching functionality, which is essential given that Granary data, the primary source for Canary-1B-v2 training, contains multilingual content where words from different languages appear within single-language datasets. Third, a unified tokenizer provides a consistent lexical space that serves as a strong foundation for downstream techniques such as phrase and word boosting.

To ensure balanced representation across all 25 languages and avoid under-representation of low-resource languages, we trained the SentencePiece tokenizer \cite{sentencepiece} on a carefully constructed dataset. The core of this training set was our En→X portion of the training data, which, as shown in Figure~\ref{fig:language-distribution}, is uniformly balanced across non-English languages. To complement this, we added English text from the YouTube subset (YTC, YODAS) of the Granary corpus as well as from NeMo ASR Set 3.0. This diverse domain selection was intended to prevent tokenizer vocabulary from overfitting to narrow-domain datasets (e.g., VoxPopuli parliamentary language).

We conducted experiments with three vocabulary sizes: 4,096, 8,192, and 16,384 tokens. These vocabularies include 1,162 special tokens specific to Canary model prompts, encompassing task-defining tokens such as \emph{<|timestamps|>}, \emph{<|notimestamps|>}, language ID tokens, and other placeholder special tokens. Our experiments consistently showed that larger vocabulary sizes yielded better downstream performance on both ASR and AST tasks. Furthermore, unified tokenizers with sizes 8,192 and 16,384 demonstrated similar compression rates, indicating an optimal balance between downstream performance and compression efficiency. Table~\ref{tab:fleurs-compression} illustrates compression rate comparisons on the FLEURS dataset, comparing two different Canary tokenizer sizes against GPT-4o\cite{openai2024gpt4ocard} tokenizer compression rates as baseline. Notably, our tokenizers achieve both improved average compression rates and substantially lower standard deviations, suggesting more consistent behavior across languages. This stability can be attributed to balancing the amount of text per language in the overall corpus supplied to SentencePiece tokenizer training, which effectively mitigates drastic cross-lingual variations in compression rates.

\begin{table}[ht]
  \centering
  \setlength{\abovecaptionskip}{10pt} % add space between table and caption
  \begin{tabular}{lccc}
    \toprule
    & GPT-4o & Canary-1B-v2 (vocab 8,192) & Canary-1B-v2 (vocab 16,384) \\
    \midrule
    Mean ($\mu$)      & 3.513 & 2.534 & 2.849 \\
    Std. Dev. ($\sigma$) & 0.662 & 0.220 & 0.230 \\
    \bottomrule
  \end{tabular}
  \caption{Compression rate comparisons on the FLEURS dataset across all 25 languages, comparing two Canary tokenizer sizes against GPT-4o as the baseline.}
  \label{tab:fleurs-compression}
\end{table}

Based on these analyses, Canary-1B-v2 employs a unified BPE tokenizer\cite{bpe-tokenizer} to handle input across all 25 languages, providing robust multilingual processing capabilities while maintaining computational efficiency.

\subsection{Pre-training}

To mimic the human learning process, which has long been observed as a multi-stage progression, we designed our pre-training to follow a two-stage process. In the first stage, the Canary model was trained on X→En (360,000 hours), English ASR (285,000 hours), and non-speech data (1,200 hours) for 150,000 steps on 64 NVIDIA A100 GPUs. The best configuration used a learning rate of 4e-4 with a minimum of 1e-6, 5,000 warm-up steps, and an inverse square-root learning rate schedule\cite{attention}. The AdamW optimizer\cite{adamw} with weight decay of 0.001 was applied. Training was initialized from a FastConformer Hybrid RNN-T/CTC checkpoint that had been trained on four languages (English, Spanish, German, and French) for ASR task. The training data was divided into duration bins using NeMo’s implementation of 2D duration bucket estimation\footnote{\url{https://github.com/NVIDIA-NeMo/NeMo/blob/main/scripts/speech_recognition/estimate_duration_bins_2d.py}}, which was then combined with Lhotse dynamic bucketing. To maximize GPU usage, OOMptimizer\cite{oomptimizer} was employed to determine the maximum feasible batch size for each bucket, and GPU utilization remained consistently high at around 95\% throughout this stage.

After completing the first stage, training continued from the resulting checkpoint on the full blend of data (ASR, X→En, and En→X), comprising roughly 1.7 million hours, for an additional 100,000 steps. A more detailed breakdown of the task composition is provided in Section~\ref{sec:data-training-data}. We observed performance saturation when training was extended beyond this point. The best results were obtained with a learning rate of 3e-4 and 5,000 warm-up steps, while other settings remained unchanged from the first stage.

In parallel, we trained a single-stage baseline model. Starting from the same FastConformer Hybrid RNN-T/CTC checkpoint, this model was trained on the full dataset from the outset for 250,000 steps, using the same optimization setup described above.

Table~\ref{tab:asr-ast-pre-training-stages} compares the two approaches. The results are close overall, with each approach slightly outperforming the other in certain cases. However, across evaluations we consistently observed that X→En scores fell short of SOTA results, and the two-stage model performed slightly better in this direction. For this reason, we selected the two-stage checkpoint for subsequent fine-tuning. An additional advantage of the two-stage setup lies in its efficiency for experimentation. In the single-stage regime, every trial with the full dataset requires training for the entire 250,000 steps, since the model must simultaneously learn both the foundational representations and the interactions between tasks. By contrast, in the two-stage regime the first stage already establishes a strong foundation. As a result, the second stage needs only around 100,000 additional steps on the full dataset to adapt and integrate task mixtures. This makes it possible to freely vary weighting schemes or data compositions in the second stage for experimentation purposes without incurring the cost of retraining the model for 250,000 steps each time.

% \paragraph{Implementation note: Parakeet-TDT-0.6B-v3.}

\emph{Parakeet-TDT-0.6B-v3} simialr to Canary-1b-v2 used FastConformer encoder with 24 layers and a TDT decoder\cite{xu2023efficient}. This model was first initialized from a multilingual CTC checkpoint pre-trained on the Granary ASR subset and trained for 150,000 steps on 128 A100 GPUs with temperature-based sampling ($\alpha = 0.5$ and $\beta = 0.5$) to balance corpora and languages, as described in Section~\ref{sec:lang-corpora-balancing}.

\begin{table}[t]
\centering
\small
\begin{threeparttable}

% (a) ASR
\begin{subtable}{\textwidth}
\centering
\begin{tabularx}{\textwidth}{l l YYYY}
\toprule
\textbf{Regime} & \textbf{Model} &
HF Leaderboard (En) & FLEURS (25) & MLS (5) & CoVoST2 (13) \\
\midrule
Two-stage & X$\to$En & 7.54\% & - & - & - \\
          &   X$\to$En + ASR + En$\to$X & 8.36\% & 11.03\% & \textbf{7.21\%} & 8.31\% \\
\midrule
Single-stage & X$\to$En + ASR + En$\to$X & \textbf{7.73\%} & \textbf{10.66\%} & 7.35\% & \textbf{8.11\%} \\
\bottomrule
\end{tabularx}
\caption{ASR results (WER). Lower is better.}
\end{subtable}

\vspace{10pt} % <-- 10pt gap

% (b) AST X->En
\begin{subtable}{\textwidth}
\centering
\begin{tabularx}{\textwidth}{l l YY}
\toprule
\textbf{Regime} & \textbf{Model} &
FLEURS (24) & CoVoST2 (11) \\
\midrule
Two-stage & X$\to$En & 70.57 & 77.25 \\
          & X$\to$En + ASR + En$\to$X & \textbf{73.23} & \textbf{75.76} \\
\midrule
Single-stage & X$\to$En + ASR + En$\to$X & 73.18 & 75.39 \\
\bottomrule
\end{tabularx}
\caption{AST (X$\to$En) results (COMET). Higher is better.}
\end{subtable}

\vspace{10pt} % <-- 10pt gap

% (c) AST En->X
\begin{subtable}{\textwidth}
\centering
\begin{tabularx}{\textwidth}{l l YY}
\toprule
\textbf{Regime} & \textbf{Model} &
FLEURS (24) & CoVoST2 (5) \\
\midrule
Two-stage & X$\to$En + ASR + En$\to$X & 83.86 & 80.14 \\
\midrule
Single-stage & X$\to$En + ASR + En$\to$X & \textbf{84.30} & \textbf{80.63} \\
\bottomrule
\end{tabularx}
\caption{AST (En$\to$X) results (COMET). Higher is better.}
\end{subtable}

\caption{Comparison of Two-stage vs. Single-stage models across ASR and AST benchmarks.}
\label{tab:asr-ast-pre-training-stages}

\begin{tablenotes}
\centering
\small
\item[$^{*}$] COMET scores computed with \href{https://huggingface.co/Unbabel/wmt22-comet-da}{\texttt{Unbabel/wmt22-comet-da}}.
\end{tablenotes}

\end{threeparttable}
\end{table}

\subsection{Fine-tuning}
\label{sec:fine-tuning}

As discussed in Section~\ref{sec:lang-coverage-imbalance}, a recurring challenge in our training pipeline is single-corpus or domain domination, which affects more than half of the supported languages. This has already impacted our results: for instance, we consistently observed unusually low (below SOTA) performance in the X→En direction, where VoxPopuli dominance is particularly evident.

To mitigate this and other imbalances in the pre-training data, we extended the pre-trained checkpoint with a a third high-quality fine-tuning stage. Importantly, this fine-tuning does not introduce a new task or domain but instead focuses the model on a high-quality subset of the training data. This subset includes NeMo ASR Set 3.0 and the YouTube portion of the Granary dataset for ASR and X→En, as well as our supplementary dataset for En→X. Additionally, we filtered the translation training data by QE scores\footnote{QE scores computed with \href{https://huggingface.co/Unbabel/wmt22-comet-da}{\texttt{Unbabel/wmt22-comet-da}}}\cite{cometoid}, keeping only samples with QE > 0.85. Thus, beyond mitigating corpus dominance, this stage also addresses quality imbalances within individual corpora.

To further balance tasks and languages, we constructed the training data as follows: for each language pair, we selected 200 hours of data from the high-quality subset. If fewer than 200 hours were available, the remainder was filled with other data (including VoxPopuli when necessary). This construction strategy was based on similar steps followed for 
\emph{parakeet-tdt-0.6b-v2}\footnote{\url{https://huggingface.co/nvidia/parakeet-tdt-0.6b-v2}}
, where it had already yielded remarkable results. The complete training set was then divided into four equally weighted groups: ASR (non-English) (4,800 h), X→En (4,800 h), En→X (4,800 h), and English ASR (600 h). During training, sampling was balanced across groups, with each group selected with equal probability. Within each group, language pairs were also equally likely to be sampled, since their data volumes were normalized to 200 hours per pair.

With this configuration, we fine-tuned for 10k additional steps (without warm-up) using inverse square-root scheduling. The best performance was achieved with a learning rate of 2e-5 and a minimum of 1e-6.

\begin{figure}[ht]
    \centering
    \includegraphics[width=0.9\textwidth]{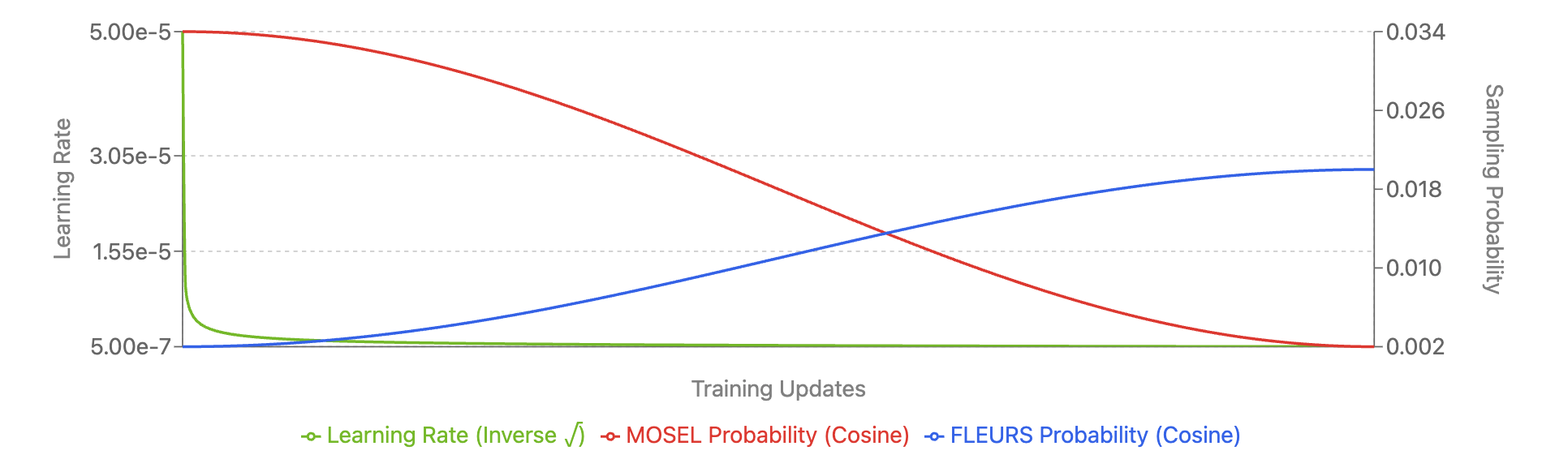}
    \caption{Example of Data Balancing Schedule. Green line shows the learning rate (inverse square root schedule), while red and blue lines represent MOSEL and FLEURS sampling probabilities (cosine schedules) along the fine-tuning process.}
    \label{fig:weight_scheduling}
\end{figure}

In addition, we explored an alternative setup inspired by \cite{parmar2024reusedontretrainrecipe}, where high-quality subsets are introduced at points in training when the learning rate is most suitable to absorb a shift in data distribution. We extended this approach by manipulating sampling weights dynamically during training. Specifically, at each dataloader step, language and corpus weights were gradually adjusted from an initial imbalance toward the target balance. This avoids “shocking” the model at the beginning of fine-tuning and instead enables a smoother adaptation. In our case, we again used the four-group configuration, balancing languages and corpora within each group as described in Section~\ref{sec:lang-corpora-balancing} (\(\alpha = 0.2, \beta = 0.5\)). Over the 10,000 fine-tuning steps, weights transitioned from the start values to the target uniform distribution (1 / number of languages in a group) according to a cosine schedule. We also experimented with linear and exponential schedules, but cosine yielded the most stable and robust results. An example of this setup is shown in Figure~\ref{fig:weight_scheduling}, where as fine-tuning progresses and the learning rate decreases, the MOSEL dataset weight gradually decreases, while the FLEURS dataset weight which is considered a high-quality dataset increases according to the predefined scheduler.

\begin{table}[ht]
\centering
\small
\begin{threeparttable}

% (a) ASR
\begin{subtable}{\textwidth}
\centering
\begin{tabularx}{\textwidth}{l YYYY}
\toprule
\textbf{Model} &
HF Leaderboard (En) & FLEURS (25) & MLS (5) & CoVoST2 (13) \\
\midrule
Pre-trained baseline & 8.36\% & 11.03\% & 7.21\% & 8.31\% \\
\midrule
Fine-tuned (15k HQ) & \textbf{7.15\%} & 8.69\% & \textbf{7.11\%} & 10.33\% \\
Fine-tuned (Balancing schedule) & \textbf{7.15\%} & \textbf{8.40\%} & 7.27\% & \textbf{8.85\%} \\
\bottomrule
\end{tabularx}
\caption{ASR results (WER). Lower is better.}
\end{subtable}

\vspace{10pt} % <-- gap

% (b) AST X->En
\begin{subtable}{\textwidth}
\centering
\begin{tabularx}{\textwidth}{l YY}
\toprule
\textbf{Model} &
FLEURS (24) & CoVoST2 (11) \\
\midrule
Pre-trained baseline & 73.23 & 75.76 \\
\midrule
Fine-tuned (15k HQ) & 79.27 & 76.82 \\
Fine-tuned (Balancing schedule) & \textbf{79.30} & \textbf{77.48} \\
\bottomrule
\end{tabularx}
\caption{AST (X$\to$En) results (COMET). Higher is better.}
\end{subtable}

\vspace{10pt} % <-- gap

% (c) AST En->X
\begin{subtable}{\textwidth}
\centering
\begin{tabularx}{\textwidth}{l YY}
\toprule
\textbf{Model} &
FLEURS (24) & CoVoST2 (5) \\
\midrule
Pre-trained baseline & 83.86 & 80.14 \\
\midrule
Fine-tuned (15k HQ) & 84.38 & \textbf{80.32} \\
Fine-tuned (Balancing schedule) & \textbf{84.56} & 80.29 \\
\bottomrule
\end{tabularx}
\caption{AST (En$\to$X) results (COMET). Higher is better.}
\end{subtable}

\caption{Comparison of pre-trained vs. fine-tuned models across ASR and AST benchmarks.}
\label{tab:asr-ast-finetuning}
\end{threeparttable}
\end{table}

Table~\ref{tab:asr-ast-finetuning} presents the results of these two setups compared to the initial pre-trained checkpoint. Fine-tuning with rebalanced data distribution successfully mitigated narrow-domain and corpus imbalance effects for ASR and X→En tasks. Improvements are particularly visible on the FLEURS dataset, which offers cleaner and more controlled evaluation conditions than CoVoST2. We achieved an ~6 absolute COMET improvement (73.23 → 79.30) for X→En, and a ~25\% relative WER reduction (11.03\% → 8.40\%) for ASR. As expected, En→X performance remained stable across setups, since the original pre-training distribution for this direction was already well-balanced.

Comparing the two fine-tuning strategies, overall performance is similar, with the exception of ASR on CoVoST2, where weight scheduling yielded a 1.5\% absolute WER gain. Since CoVoST2 is the most spontaneous of our benchmarks, we hypothesize that weight scheduling prevents “catastrophic forgetting” of pre-trained knowledge, whereas fine-tuning only on 15k hours of clean data biases the model too strongly toward clearer speech. Moreover, weight scheduling consistently extracted more performance from the available data, leading to additional gains in the X→En direction.

In light of these findings, we release the weight-scheduled fine-tuned model as our primary version and continue to explore data balancing schedules to better understand their benefits and broader applicability.

% \paragraph{Implementation note: Parakeet-TDT-0.6B-v3.}

Fine-tuning stage for \emph{Parakeet-TDT-0.6B-v3} ran for 5,000 steps on 4 A100 GPUs using 7,500 hours of high-quality NeMo ASR Set 3.0 data.

\section{Timestamps}
% \section{Timestamps}
\label{sec:timestamps}

\subsection{Timestamps Generation in Attention-Encoder Decoder Models}

The extraction of precise timestamps from spoken words is a critical capability of modern Speech Recognition systems, essential for powering downstream applications such as timed subtitling and content retrieval. The methods for extracting timestamps from attention-based speech recognition models are diverse, ranging from post-processing techniques to architectural redesigns. Each approach represents a different trade-off between accuracy, computational overhead, and architectural complexity.

The architecture that we are working with in Canary-1B-v2 is an attention-based encoder-decoder (AED) system. In these models, alignment between audio frames and text tokens is handled during the decoding phase. This is primarily facilitated by the cross-attention mechanism, which produces a weight matrix representing the probabilistic correlation between each output token and every input audio frame. While these weights provide a "soft" alignment, they are prone to non-monotonicity, where the attention path can jump back and forth in time, leading to a certain degree of inherent unreliability or "fuzziness."

\subsubsection{Post-Processing with Dynamic Time Warping (DTW)}
To convert this soft, often non-monotonic alignment into precise, word-level boundaries, a robust post-processing algorithm like Dynamic Time Warping (DTW)\cite{DTWAlignment} is applied to the cross-attention matrix. DTW finds the optimal, non-linear alignment path between two sequences by computing a cumulative cost or score matrix. In this context, it aligns the sequence of output tokens with the sequence of input audio frames by finding the path of highest cumulative attention scores. This process effectively "hardens" the fuzzy attention into a clear, monotonic sequence of word boundaries, providing the start and end times for each word. A prominent example of this method is its use in third-party implementations\cite{lintoai2023whispertimestamped} for Whisper\cite{radford2023whisper} model, which natively provides only segment-level timestamps.

\subsubsection{Predictive Timestamp Generation}

A fundamentally different approach is to embed timestamp prediction directly into the AED model itself. In this setup, timestamping is reframed as a core predictive task rather than a post-processing step. Whisper \cite{radford2023whisper} first introduced this strategy by training on a corpus in which special frame-number tokens (e.g., \emph{<|0|>, <|1|>, …}) were inserted into transcriptions at segment boundaries. The Canary model \cite{canary1b-flash-timestamps} subsequently adopted and extended this approach by inserting the tokens at word boundaries. By learning to predict these tokens directly, the model incorporates an understanding of acoustic–temporal relationships into its core predictive capability.

This approach has distinct advantages, as it shifts the alignment burden to the model's training process. However, it also introduces new complexities:

\begin{itemize}
    \item \textbf{Data Curation:} The creation of a diverse and well-represented training corpus with accurate timestamp tokens requires significant computational resources and careful data curation.
    \item \textbf{Training Challenges:} The model must now learn an additional task, which can create balancing issues and requires a more carefully tuned training regimen to avoid negatively impacting transcription accuracy.
\end{itemize}

\subsubsection{Forced Alignment as a Standalone Method}

For many production and commercial applications\cite{whisperX}, the process of timestamping has been abstracted away into a service. Forced alignment, a method that takes an audio file and its corresponding transcript to generate timestamps, serves as a powerful standalone. 
Given the complexities and potential instability of relying solely on "soft" attention weights or multi-task learning for timestamps, we explored exactly the forced alignment method as a standalone approach for Canary-1B-v2. Unlike the methods above, it does not rely on the end-to-end model's internal alignment mechanism for timestamping.

For our purposes, the NeMo Forced Aligner (NFA)\cite{nfa} has proven to be an effective method for generating word and segment-level timestamps in \cite{canary1b-flash-timestamps}. This method has become the backbone for timestamp extraction in our system, used in conjunction with auxiliary Multilingual Parakeet CTC model.

\subsection{NeMo Forced Aligner (NFA)}

NeMo Forced Aligner (NFA) is a tool for generating token-, word-, and segment-level timestamps of speech in audio using CTC-based ASR models. Optionally, NFA can align ground truth text with the CTC-based ASR model decoder's log-probabilities. The alignment process uses Viterbi decoding to find the most probable alignment of given tokens based on log-probabilities from the CTC head.

\begin{figure}[htbp]
  \centering
  % Subfigure (a) on top
  \begin{subfigure}{\textwidth}
    \centering
    \includegraphics[width=0.95\linewidth]{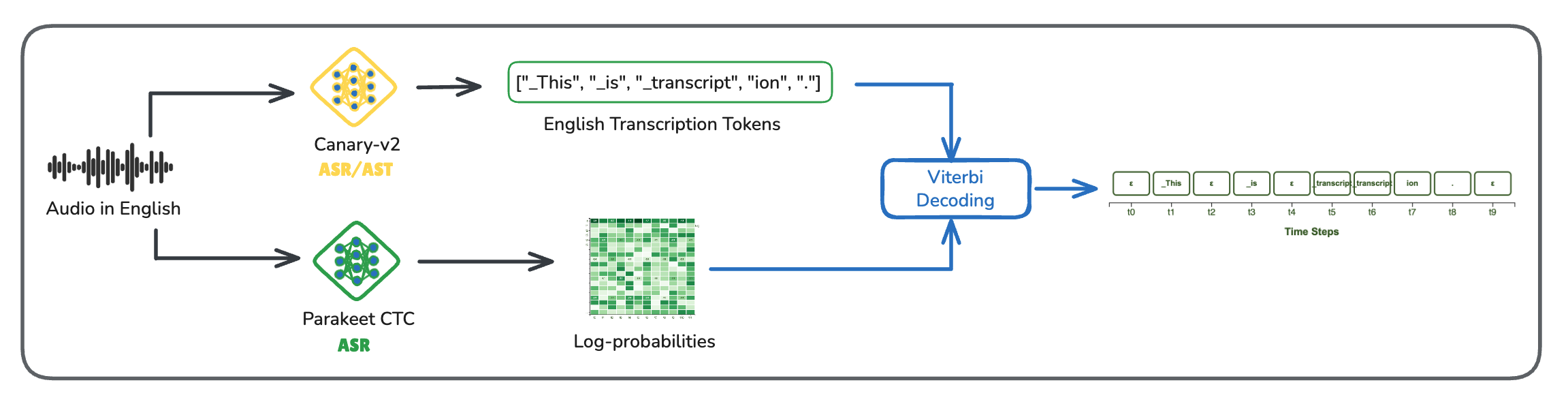}
    \caption{Timestamps for ASR output.}
    \label{fig:canary-timestamps-asr}
  \end{subfigure}

  \vspace{1em} % optional vertical space between (a) and (b)

  % Subfigure (b) below
  \begin{subfigure}{\textwidth}
    \centering
    \includegraphics[width=0.95\linewidth]{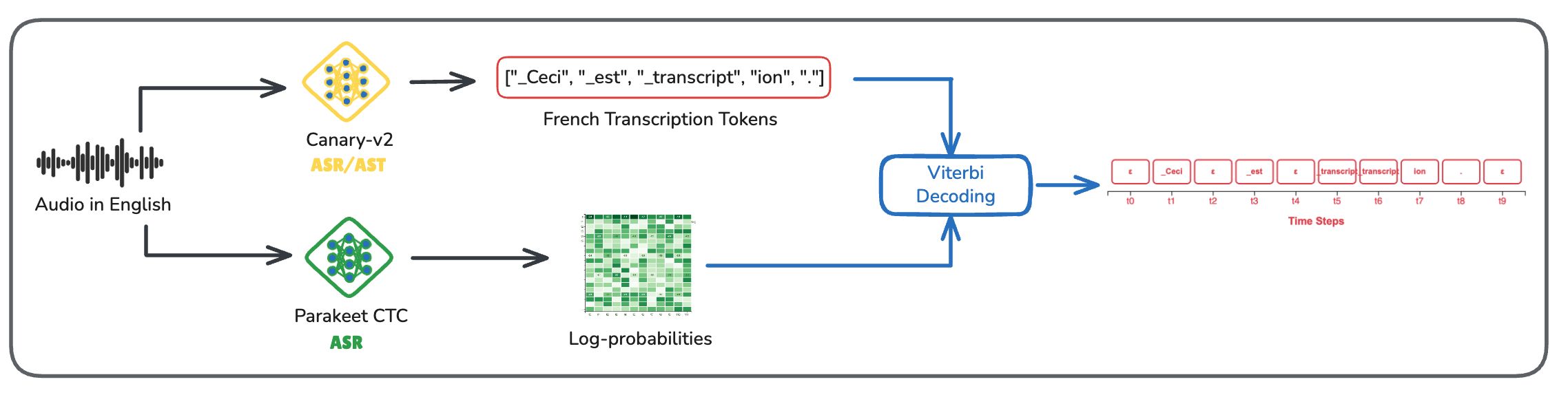}
    \caption{Timestamps for AST output.}
    \label{fig:canary-ast}
  \end{subfigure}

  \caption{Overview of Timestamps Generation for Canary-1B-v2 with NFA.}
  \label{fig:canary-timestamps-ast}
\end{figure}

Figure~\ref{fig:canary-timestamps-asr} depicts the timestamp generation process for Canary-1B-v2 speech recognition output using NFA capabilities. This pipeline incorporates an additional ASR model, a Parakeet architecture model with CTC decoder comprising 600M parameters. This model was trained using the same unified tokenizer as Canary-1B-v2 for 250,000 steps on 128 NVIDIA A100 GPUs, using the ASR subset of Canary-1B-v2 training data described in Section~\ref{sec:data}.
When integrating the Forced Alignment pipeline into Canary-1B-v2's inference pipeline, a critical question emerged regarding ASR and AST task alignment: Can NFA effectively align translated speech with audio sequences in completely different languages?

ASR alignment is typically monotonic, maintaining a strictly sequential, left-to-right mapping from temporal audio frames to output phonemes, characters, or words. This "forced alignment" ensures linear correspondence between transcription output and input speech signal.
In contrast, AST alignment is both non-monotonic and cross-lingual. Words and phrases in the source language lack direct, sequential correspondence with translated target language output due to linguistic phenomena including different grammatical structures, verb tenses, and word orders. For example, the English phrase \emph{I love you} translates to French \emph{je t'aime}, where source and target word orders don't align sequentially. 

Additionally, the CTC model (Parakeet CTC) incorporated into the NFA pipeline is trained solely for ASR tasks and lacks AST knowledge.

\begin{figure}[htbp]
  \centering
  \includegraphics[width=0.6\linewidth]{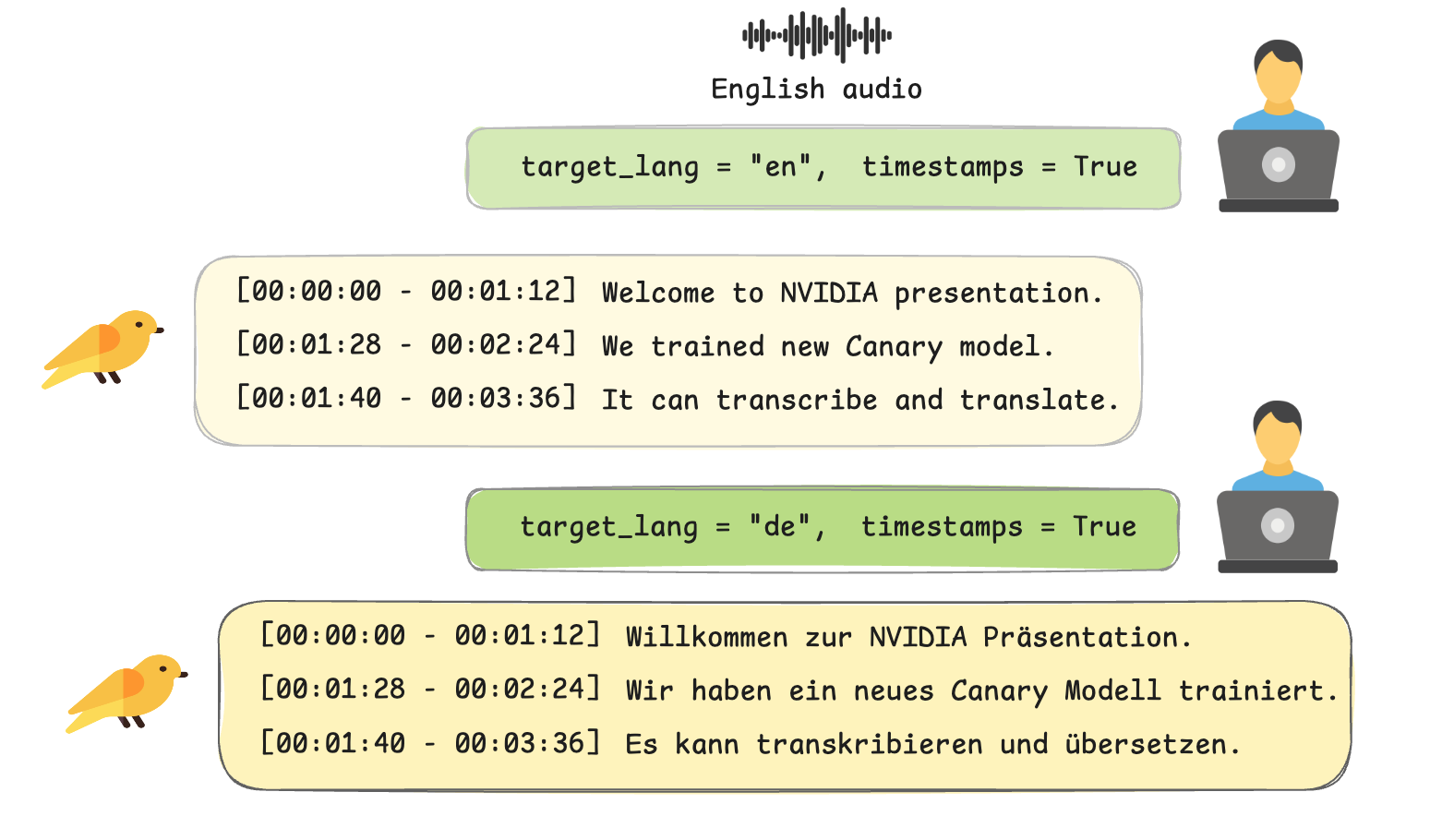}
  \caption{Examples of Canary-1B-v2 output along with segment-level timestamps.}
  \label{fig:output-examples}
\end{figure}

To evaluate timestamp generation capabilities for AST tasks, we conducted experiments on evaluation benchmarks and performed manual evaluation of the generated timestamps. The NFA pipeline integrated into Canary-1B-v2's inference mechanism successfully produced high-quality segment-level timestamps, as shown in Figure~\ref{fig:canary-timestamps-ast}b. We therefore recommend using segment-level timestamps with translation outputs, as word-level timestamps can be inaccurate due to the non-monotonic nature of speech translation. An example of segment-level timestamps with the respective prompts is illustrated in Figure~\ref{fig:output-examples}. 

Overall, we observed that the current pipeline with NFA yields consistently reliable timestamping performance. We attribute this in part to the fact that our evaluation scope primarily covers European languages, where translations to and from English tend to preserve sentence structure without drastic reordering. By contrast, languages with more divergent syntactic structures such as Mandarin may introduce substantial shifts in word and phrase order. As such, while the results within our explored language set are promising, the applicability of the NFA pipeline to typologically distant languages remains uncertain and warrants more thorough testing and investigation.

Looking ahead, future work will focus on more comprehensive and quantitative AST timestamp evaluation and on exploring how log-probabilities of an ASR model are effectively leveraged for aligning translation transcript tokens.

\section{Evaluation}
% \section{Evaluation}
\label{sec:eval}

The evaluation of the final \emph{Canary-1B-v2} model is conducted using the benchmark suites introduced in Section~\ref{sec:evaluation-benchmarks}. For comparative analysis, we consider several strong baselines: \emph{Whisper-large-v3} (1.55B), \emph{seamless-m4t-v2-large} (2.3B), and \emph{seamless-m4t-medium} (1.2B), as they are the closest in scale to \emph{Canary-1B-v2} among seamless speech models. In addition, we include \emph{Voxtral-Mini-3B-2507}\cite{liu2025voxtral} (3B) and \emph{Phi-4-multimodal-instruct}\cite{abouelenin2025phi4multimodal} (5.6B), two multimodal and LLM-based systems, in order to highlight the trade-offs between accuracy and inference efficiency where \emph{Canary-1B-v2} provides an advantage. Our second release, \emph{Parakeet-TDT-0.6B-v3}, is likewise included in the multilingual ASR evaluations.

Multilingual evaluations in this section are reported under two complementary settings:  
\begin{enumerate}
    \item \textbf{All supported languages:} 24 total, excluding Latvian, which is not supported by \emph{seamless-m4t-v2-large} or \emph{seamless-m4t-medium}.  
    \item \textbf{Common languages:} 6 shared by all compared models (\emph{en}, \emph{fr}, \emph{de}, \emph{it}, \emph{pt}, and \emph{es}).  
\end{enumerate}

\subsection{ASR Performance}

For ASR performance evaluation, we report normalized WER (\%). We use the Hugging Face Open Automatic Speech Recognition Leaderboard normalizers\footnote{\url{https://github.com/huggingface/open_asr_leaderboard/tree/main/normalizer}}: the English normalizer for English test sets, and the newly added multilingual normalizer for non-English test sets. Both remove punctuation and lowercase the text. The English normalizer further performs richer, language-specific normalization (e.g., numbers, abbreviations, currencies), whereas the multilingual normalizer emphasizes cross-language consistency by replacing symbols and punctuation-like markers with spaces and by mapping or stripping diacritics to their base forms.

\subsubsection{English ASR (HuggingFace Leaderboard) Performance}

Table~\ref{tab:asr-performance} presents the results for the Hugging Face Leaderboard datasets, calculated with the official repository and reported on the Hugging Face Open ASR Leaderboard. Notably, \emph{Canary-1B-v2} outperforms \emph{Whisper-large-v3} on the official average WER metric, while also achieving remarkable inference efficiency. With an RTFx of 749, \emph{Canary-1B-v2} runs approximately 7--10 times faster than the other evaluated models, highlighting its strength in balancing accuracy with throughput efficiency.

Additionally, \emph{Parakeet-TDT-0.6B-v3} attains the highest throughput among all systems (RTFx \(=3332.74\)) while maintaining a low average WER of \(6.32\%\), within \(0.18\) absolute points of \emph{Phi-4-multimodal-instruct} (\(6.14\%\)). This corresponds to \(\approx 54\times\) faster inference than \emph{Phi-4-multimodal-instruct}, combining near–state-of-the-art accuracy with extreme throughput.

\begin{table}[ht]
\centering
\resizebox{\textwidth}{!}{%
\begin{tabular}{lcccccccccc}
\toprule
\textbf{Model} & \textbf{RTFx} & \textbf{Avg WER} & \textbf{AMI} & \textbf{Earnings22} & \textbf{Gigaspeech} & \textbf{LS Clean} & \textbf{LS Other} & \textbf{SPGI} & \textbf{TED-LIUM} & \textbf{VoxPopuli} \\
\midrule
\emph{Whisper-large-v3}          & 145.51 & 7.44 & 15.95 & 11.29 & 10.02 & 2.01 & 3.91 & 2.94 & 3.86 & 9.54 \\
\emph{Voxtral-Mini-3B-2507}      & 109.86 & 7.05 & 16.31 & 10.65 & 10.24 & 1.89 & 4.08 & 2.37 & 3.70 & 7.14 \\
\emph{Phi-4-multimodal-instruct} &  62.12 & 6.14 & 11.45 & 10.50 &  9.77 & 1.67 & 3.82 & 3.11 & 2.89 & 5.93 \\
\midrule
\emph{Parakeet-TDT-0.6B-v3}              & 3332.74 & 6.32 & 11.39 & 11.19 & 9.57 & 1.92 & 3.59 & 3.98 & 2.8 & 6.09 \\
\midrule
\emph{Canary-1B-v2}              & 749.00 & 7.15 & 16.01 & 11.79 & 10.82 & 2.18 & 3.56 & 2.28 & 4.29 & 6.25 \\
\bottomrule
\end{tabular}%
}
\vspace{10pt}
\caption{ASR performance (WER \%) and inference efficiency (RTFx) on Hugging Face OpenASR Leaderboard benchmarks.}
\label{tab:asr-performance}
\end{table}

\subsubsection{Multilingual ASR Performance}

\begin{figure}[htbp]
    \centering
    \includegraphics[width=0.99\linewidth]{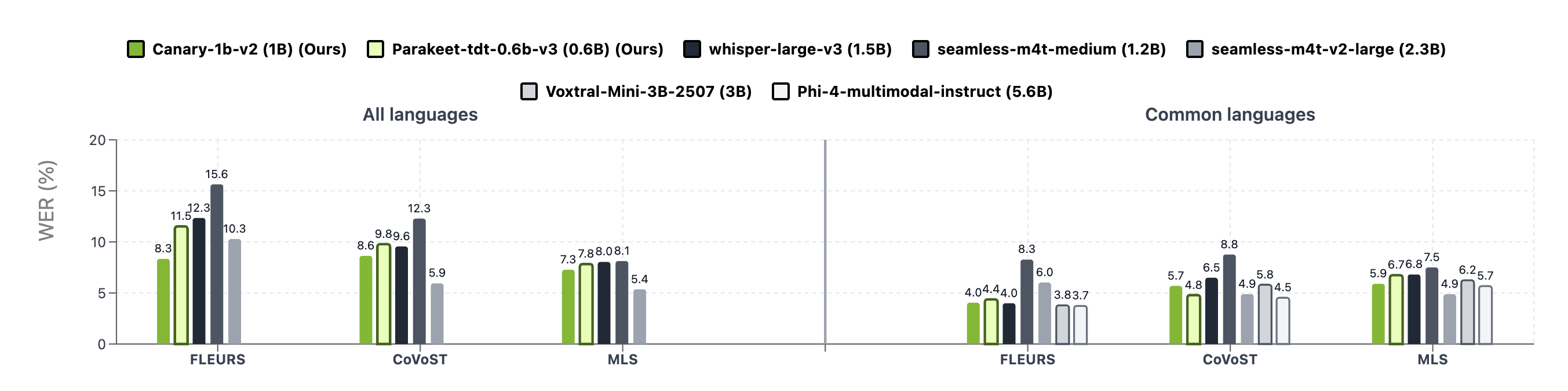}
    \caption{Multilingual ASR performance of \emph{Canary-1B-v2} and \emph{Parakeet-TDT-0.6B-v3} compared to baseline models on both 24-language and 6-language subsets.}
    \label{fig:multilingual-asr}
\end{figure}

As is seen in Figure~\ref{fig:multilingual-asr}, \emph{Canary-1B-v2} demonstrates consistently strong multilingual ASR performance across both the full 24-language set and the common 6-language subset. On the full evaluation, computed as an average across three different datasets (FLEURS, CoVoST, MLS), it achieves an average WER of around 8.1\%, outperforming \emph{whisper-large-v3} (9.9\%) and \emph{seamless-m4t-medium} (12.0\%), while remaining only slightly behind the much larger and heavier \emph{seamless-m4t-v2-large} (7.2\%). On the common-language evaluation, also averaged across the same three datasets, \emph{Canary-1B-v2} reaches 5.2\% WER compared to 5.8\% for \emph{whisper-large-v3} and 8.2\% for \emph{seamless-m4t-medium}, and in fact slightly outperforms \emph{seamless-m4t-v2-large} (5.3\%). Even though other models are specialized in this smaller set, \emph{Canary-1B-v2} supports 19 additional languages while still matching or surpassing much heavier and LLM-powered baselines.

Additionally, the companion 0.6B model \emph{Parakeet-TDT-0.6B-v3} averages 9.7\% WER across the 24-language evaluation (11.52/9.78/7.83 on FLEURS/CoVoST/MLS), edging past \emph{whisper-large-v3} (9.9\%) while trailing \emph{Canary-1B-v2} (8.1\%) and \emph{seamless-m4t-v2-large} (7.2\%). On the common-language subset it reaches 5.3\% WER (4.37/4.79/6.74), essentially matching \emph{seamless-m4t-v2-large} (5.3\%) and outperforming \emph{whisper-large-v3} (5.8\%).

Per-language WERs for each evaluation benchmark are provided in the Appendix~\ref{app:eval_details_asr}

\subsubsection{Noise-robustness evaluation}

Table~\ref{tab:noise-robustness} reports the robustness of \emph{Canary-1B-v2} and \emph{Parakeet-TDT-0.6B-v3} across different Signal-to-Noise Ratios (SNR) on the LibriSpeech Clean test set\cite{librispeech}, where MUSAN music and noise samples~\cite{snyder2015musanmusicspeechnoise} were added.

Interestingly, we observe that performance of \emph{Canary-1B-v2} slightly improves when decreasing the SNR from 100 to 25, with WER dropping from 2.18\% to 2.01\%. This suggests that \emph{Canary-1B-v2} benefits from a small amount of background noise, likely due to training on large-scale datasets that naturally include noisy conditions, such as the YouTube subset of the Granary corpus and VoxPopuli, where background noise from live interpreters and other sources is present. This indicates that the model has learned to leverage noisy input scenarios, resulting in improved robustness.  

Complementing this, \emph{Parakeet-TDT-0.6B-v3} is consistently more robust across SNRs (1.92–1.96\% WER from 100 to 25 dB) and outperforming under severe noise with 12.21\% at \(-5\) dB vs 19.38\% for \emph{Canary-1B-v2} (\(\sim\)37\% relative).

\begin{table}[ht]
\centering
\begin{tabular}{lccccccc}
\toprule
\ & \multicolumn{7}{c}{\textbf{SNR (dB)}} \\
\cmidrule(lr){2-8}
\textbf{Model} & \textbf{100} & \textbf{50} & \textbf{25} & \textbf{10} & \textbf{5} & \textbf{0} & \textbf{-5} \\
\midrule
\emph{Parakeet-TDT-0.6B-v3} & \textbf{1.92} & \textbf{1.92} & 1.96 & 2.15 & 2.62 & 4.82 & 12.21 \\
\midrule
\emph{Canary-1B-v2} & 2.18 & 2.16 & \textbf{2.01} & 2.29 & 2.80 & 5.08 & 19.38 \\
\bottomrule
\end{tabular}

\vspace{10pt}

\caption{WER (\%) across different SNR levels on the LibriSpeech Clean test set with MUSAN music and noise samples.}
\label{tab:noise-robustness}
\end{table}

\subsection{AST Performance}

For AST performance we consider two evaluation directions: X$\to$En En and En $\rightarrow X$. We again used the evaluation benchmarks introduced in Section~\ref{sec:evaluation-benchmarks}.

In terms of performance metrics, we report both COMET and BLEU\cite{papineni2002bleu} scores. COMET is used as the primary metric throughout this section, while BLEU scores are reported in Appendix~\ref{app:eval_details_ast_x-en} and \ref{app:eval_details_ast_en-x}. BLEU, being based on $n$-gram overlap, often underestimates semantic adequacy when translations diverge lexically or syntactically from reference sentences but remain valid. This makes BLEU less reliable, especially in multilingual and low-resource settings. COMET, in contrast, leverages pretrained multilingual encoders and human-annotated data, enabling it to better capture semantic similarity and translation quality beyond surface-level word overlap. As such, COMET provides a more faithful reflection of translation performance, while BLEU is included for completeness and comparability with prior work.

\subsubsection{\texorpdfstring{X$\to$En Performance}{X→En Performance}}

\begin{figure}[htbp]
    \centering
    \includegraphics[width=0.99\linewidth]{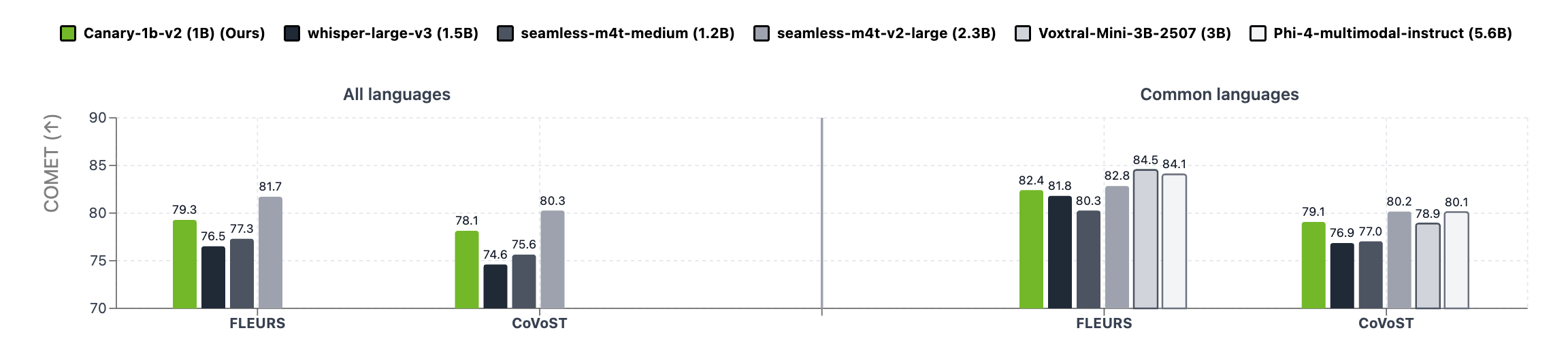}
    \caption{AST (X$\to$En) performance of \emph{Canary-1B-v2} compared to baseline models on both 24-language and 6-language subsets.}
    \label{fig:ast-x-en}
\end{figure}

As shown in Figure~\ref{fig:ast-x-en}, \emph{Canary-1B-v2} outperforms \emph{whisper-large-v3} and \emph{seamless-m4t-medium} by a large margin across both evaluation benchmarks. On the full-language setting, it achieves 79.28 COMET on FLEURS and 78.14 on CoVoST2, compared to 76.51 / 74.61 for \emph{whisper-large-v3} and 77.30 / 75.64 for \emph{seamless-m4t-medium}. Even against the much larger \emph{seamless-m4t-v2-large} (81.71 / 80.26), \emph{Canary-1B-v2} remains close despite having less than half the parameters.  

On the common-language evaluation, \emph{Canary-1B-v2} continues to perform strongly, reaching 82.41 COMET on FLEURS and 79.07 on CoVoST. This is competitive with the heavier \emph{seamless-m4t-v2-large} (82.84 / 80.16) and even the LLM-powered \emph{Voxtral-Mini-3B-2507} (84.53 / 78.90) and \emph{Phi-4-multimodal-instruct} (84.06 / 80.11). Notably, on the more spontaneous and noisy CoVoST2 dataset, \emph{Canary-1B-v2} performs particularly well (79.07 vs.\ 76.86 for \emph{whisper-large-v3}), increasing the gap over the \emph{whisper} model, comparable to the \emph{Phi-4-multimodal-instruct} (80.11), and even outperforming the \emph{Voxtral-Mini-3B-2507} (78.90). This reflects the robustness learned from large-scale noisy training sources.

\subsubsection{\texorpdfstring{En$\to$X Performance}{En→X Performance}}

\begin{figure}[htbp]
    \centering
    \includegraphics[width=0.99\linewidth]{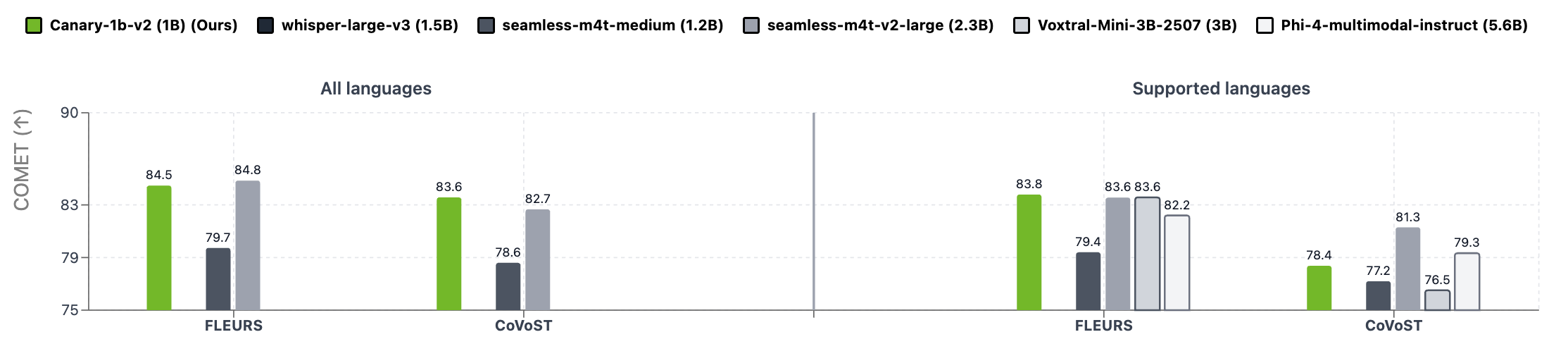}
    \caption{AST (En$\to$X) performance of \emph{Canary-1B-v2} compared to baseline models on both 24-language and 6-language subsets.}
    \label{fig:ast-en-x}
\end{figure}

As shown in Figure~\ref{fig:ast-en-x}, \emph{Canary-1B-v2} achieves strong En$\to$X performance. On FLEURS, it remains on par with the much larger \emph{seamless-m4t-v2-large} in the all-language setting (84.47 vs.\ 84.85 COMET) and even slightly outperforms all compared models in the supported-language evaluation (83.79 vs.\ 83.56 for \emph{seamless-m4t-v2-large}, 82.20 for \emph{Phi-4}, and 83.56 for \emph{Voxtral-Mini-3B-2507}). On CoVoST2, however, \emph{Canary-1B-v2} lags behind \emph{seamless-m4t-v2-large} (80.03 vs.\ 82.66 in the all-language setting, and 78.37 vs.\ 81.30 in the supported-language evaluation). A possible explanation is that while the En$\to$X training data covers diverse domains, the X-language ASR pathways were more strongly shaped by narrower-domain data, leading the decoder to generalize less effectively to spontaneous CoVoST speech.

\subsection{nGPT and FastConformer comparison}
To better understand the capabilities of nGPT as an encoder, we trained models of different scales on the Granary datasets. These experiments were designed to assess how well nGPT scales and how it compares to the FastConformer baseline across ASR and AST benchmarks.

As shown in Table \ref{tab:asr-ast-fc-ngpt}, scaling from 1B to 3B parameters consistently improved performance. The 3B-parameter nGPT model demonstrated stronger generalization across tasks, with especially notable gains on the challenging X→EN translation benchmark—where FastConformer struggled most. Importantly, the 3B model achieved these results within a single training stage and only 100k steps. While the 1B model falls short of the 3B across tasks, it still shows the same generalization trend and remains competitive, particularly on translation.

In contrast, FastConformer required a more elaborate multi-stage training regime to achieve strong results. With sufficient fine-tuning, however, it was able to surpass nGPT in all tasks. This highlights a key distinction: nGPT is a data-hungry model that thrives under large-scale training and quickly reaches competitive performance, whereas FastConformer benefits more from fine-tuning and exhibits greater relative improvements in low-data scenarios.

\begin{table}[t]
\centering
\small
\begin{threeparttable}

% (a) ASR
\begin{subtable}{\textwidth}
\centering
\begin{tabularx}{\textwidth}{l l YYYY}
\toprule
\textbf{Model} & \textbf{Regime} &
HF Leaderboard (En) & FLEURS (25) & MLS (5) & CoVoST2 (13) \\
\midrule
FC 1B   & 1st stage        & 7.73\% & 10.66\% & 7.35\% & 8.11\% \\
nGPT 1B & 1st stage        & 8.94\% &  9.94\% & 8.11\% &  9.96\% \\
nGPT 3B & 1st stage        & 7.73\% & 10.18\% & 8.35\% & 10.48\% \\
\midrule

FC 1B   & 3-stage (15k HQ) & 7.15\% & 8.69\% & 7.11\% & 10.33\% \\
nGPT 3B & 2-stage (15k HQ) & 7.32\% & 10.31\% & 8.32\% & 10.30\% \\
\bottomrule
\end{tabularx}
\caption{ASR results (WER). Lower is better.}
\end{subtable}

\vspace{10pt}

% (b) AST X->En
\begin{subtable}{\textwidth}
\centering
\begin{tabularx}{\textwidth}{l l YY}
\toprule
\textbf{Model} & \textbf{Regime} &
FLEURS (24) & CoVoST2 (11) \\
\midrule
FC   1B   & 1st stage        & 73.18 & 75.39 \\
nGPT 1B & 1st stage        & 76.72 & 74.02 \\
nGPT 3B & 1st stage        & 78.36 & 75.75 \\
\midrule
FC   1B   & 3-stage (15k HQ) & 79.27 & 76.82 \\
nGPT 3B & 2-stage (15k HQ) & 79.14 & 75.90 \\
\bottomrule
\end{tabularx}
\caption{AST (X$\to$En) results (COMET). Higher is better.}
\end{subtable}

\vspace{10pt}

% (c) AST En->X
\begin{subtable}{\textwidth}
\centering
\begin{tabularx}{\textwidth}{l l YY}
\toprule
\textbf{Model} & \textbf{Regime} &
FLEURS (24) & CoVoST2 (5) \\
\midrule
FC 1B   & 1st stage        & 84.30 & 80.63 \\
nGPT 1B & 1st stage        & 83.05 & 78.31 \\
nGPT 3B & 1st stage        & 84.22 & 80.00 \\
\midrule
FC 1B   & 3-stage (15k HQ) & 84.38 & 80.32 \\
nGPT 3B & 2-stage (15k HQ) & 84.28 & 79.48 \\
\bottomrule
\end{tabularx}
\caption{AST (En$\to$X) results (COMET). Higher is better.}
\end{subtable}

\caption{Comparison of FC and nGPT across training regimes on ASR and AST benchmarks.}
\label{tab:asr-ast-fc-ngpt}
\begin{tablenotes}
\centering
\small
\item[$^{*}$] COMET scores computed with \texttt{Unbabel/wmt22-comet-da}.
\end{tablenotes}

\end{threeparttable}
\end{table}

\subsection{Longform inference}
Neural encoders are typically trained on relatively short input sequences, which means they often fail to generalize well when faced with significantly longer contexts at inference time. For ASR, this limitation becomes especially clear when transcribing utterances beyond the average training length (e.g., >40 seconds). As context length grows, models exhibit degraded stability and accuracy, making it essential to explore strategies that overcome this bottleneck.

One class of solutions modifies the inference procedure rather than the model architecture itself. A common approach is local attention \cite{rekesh2023fast}, where each query token attends only to a fixed-size window of neighboring tokens instead of the entire sequence. This sliding-window design reduces quadratic attention costs and stabilizes inference on long inputs, while still capturing the local dependencies most critical for recognition tasks. In practice, such methods can be readily applied in models with CTC or RNN-T heads, making them an effective strategy for scaling ASR systems to longer contexts. The other method is chunking mechanism in which audio is divided into smaller segments that are transcribed separately and then merged and it can easily be adapted to encoder–decoder architectures.

\subsubsection{Chunk based approach}

For our current FastConformer encoder paired with a Transformer decoder, we employed a dynamic parallel chunking mechanism to enable efficient long-form inference. Each input audio file is segmented into chunks of 30–40 seconds, where the exact length is chosen dynamically to minimize padding in the final chunk. To preserve continuity across boundaries, adjacent chunks overlap by 1 second.

The chunked segments are then fed into the model in parallel as separate entries within a batch. After the model produces transcriptions for all segments, the outputs are merged into a single hypothesis. To resolve overlapping regions, we apply a Longest Common Subsequence (LCS) algorithm at the token level, ensuring that duplicated text in the overlap regions is removed from the final transcript.

For very long recordings (longer than one hour), we further segment the audio into hour-long blocks, each processed independently using the same chunking and merging procedure. This hierarchical design ensures scalability to arbitrarily long audio while maintaining both efficiency and transcription quality.
% put this in the preamble

   % for spanning rows/cols
We compared this method with chunking inference without overlap, where fixed audio chunks were processed sequentially through the model one by one, and observed clear performance gains. Table~\ref{tab:chunking_results} summarizes the results. Parallel chunking achieves lower WER and significantly higher RTFx on both benchmarks compared to script-based sequential chunking. On the Earnings22 dataset, WER improved from 15.61\% to 13.93\%, while RTFx increased almost 4×. Similarly, on the TAL(10h) dataset, WER dropped from 16.62\% to 10.12\%, and RTFx more than doubled.

\begin{table}[ht]
\centering
\caption{Comparison of two chunking mechanisms for long-form inference. In parallel chunking(ours), audio is split into overlapping chunks that are processed simultaneously within a batch, with overlaps later merged to maintain continuity. In \textit{sequential chunking}, fixed-length chunks are processed one by one without overlap, and outputs are concatenated directly.}
\begin{tabular}{|l|c c|c c|}
\hline
 & \multicolumn{2}{c|}{\textbf{ASR Earnings22}} & \multicolumn{2}{c|}{\textbf{ASR TAL (10h)}} \\
\cline{2-5}
\textbf{Method} & EN (\%) & RTFx & EN (\%) & RTFx \\
\hline
\rowcolor{green!15}
Parallel Chunking (Ours)             & \textbf{13.93} & \textbf{144.244} & \textbf{10.12} & \textbf{64.067} \\
Sequential Chunking (30 sec) & 15.61          & 37.17            & 16.62         & 26.88 \\
\hline
\end{tabular}
\label{tab:chunking_results}
\end{table}

\subsubsection{Postional encodings for longform inference}
While inference-time strategies such as local attention and chunking mitigate some of the difficulties of long-context transcription, they do not fully address the underlying limitations of how models represent positional information. To complement these approaches, we explored alternative positional encodings within the nGPT-based encoder, focusing on Rotary Position Embeddings (RoPE) and Attention with Linear Biases (ALiBi), and further adapting them to better suit long-form ASR.

\begin{figure}[htbp]
    \centering
    \includegraphics[width=0.8\textwidth]{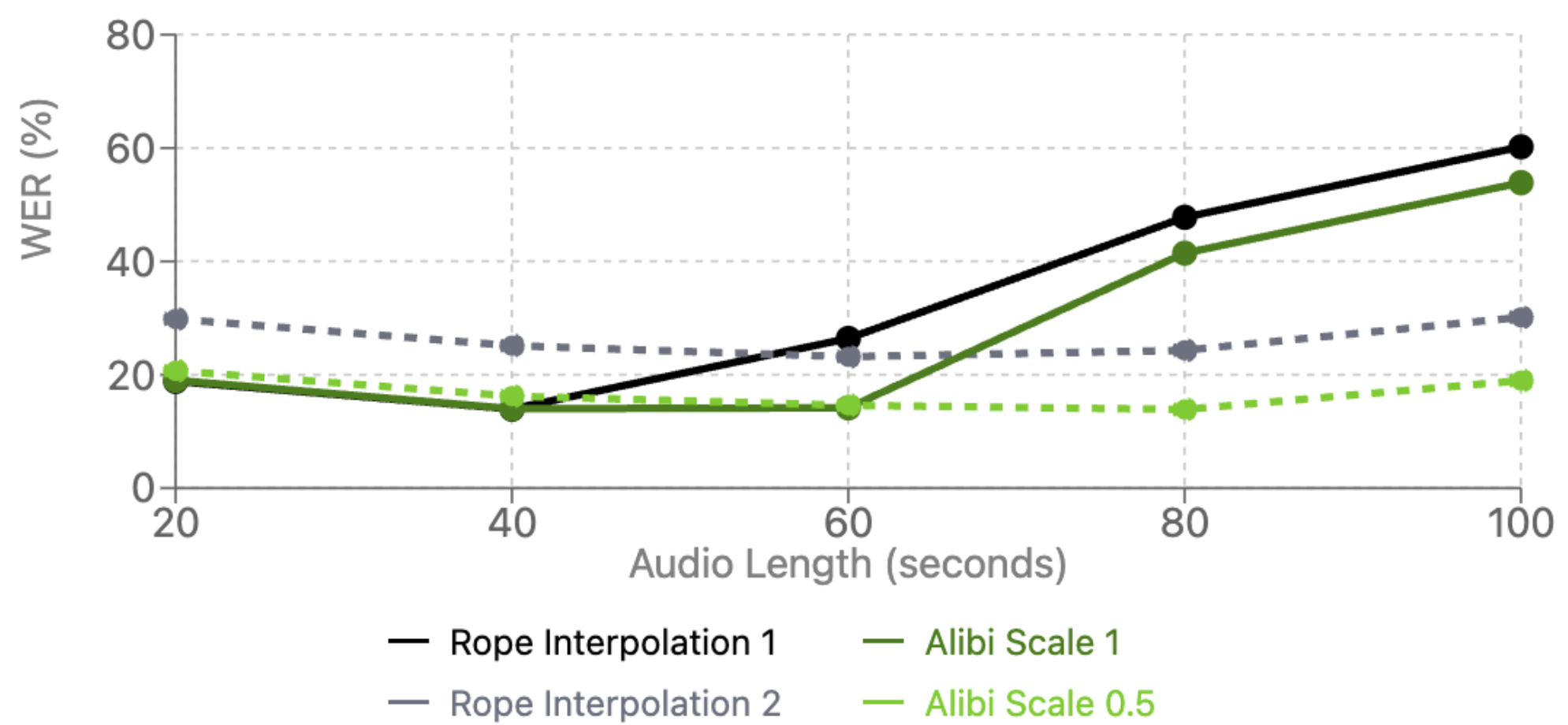}
   \caption{Comparison of Attention with Linear Biases (ALiBi) and Rotary Position Embeddings (RoPE) on long-form ASR with the Earnings dataset. For RoPE, larger interpolation factors correspond to reducing the angular rotation step to better accommodate longer sequences. For ALiBi, smaller scaling factors apply a lighter penalty to distant tokens, allowing the model to attend further into the context.}
    \label{fig:alibi_rope}
\end{figure}

\ref{fig:alibi_rope} shows Word Error Rate (WER) on the Earnings dataset across different evaluation lengths (20, 40, 60, 80, and 100 seconds). We observe that in the baseline implementations, ALiBi consistently outperforms RoPE for longer sequences, although both exhibit increasing WER as the context length grows. 

\vspace{0.3em} % space before figure
To improve generalization for nGPT encoder architecture, we modified each positional encoding. For RoPE, we reduced the angular rotation applied at each position so that longer input sequences would span the same effective angular spectrum as those seen during training. For ALiBi, we adjusted the bias scaling factor, which controls how strongly distant tokens are down-weighted, thereby allowing the model to retain more attention to faraway context. These adjustments significantly reduced WER across all lengths. Nevertheless, ALiBi remained more effective than RoPE in long-context ASR, highlighting the benefit of incorporating distance-aware attention biases in encoder-style speech tasks.

Empirically, the two approaches demonstrated task-dependent behavior. In ASR, ALiBi showed improvements, likely because reducing emphasis on distant tokens is not detrimental for speech recognition, where local context dominates. In contrast, for AST, long-range context is more critical, and we found RoPE to be more effective. These results highlight that positional encoding strategies interact differently with downstream tasks: ALiBi favors efficiency in short-to-mid context alignment, whereas RoPE better supports tasks requiring global context retention.

\section{Conclusions}
This report introduces \textit{\textbf{Canary-1B-v2}} and \textbf{\textit{Parakeet-TDT-0.6B-v3}}, two multilingual models that achieve state-of-the-art ASR and AST performance with high efficiency. The models were developed through a comprehensive exploration of data, training strategies, and architecture. This included curating a 1.7M-hour dataset with non-speech audio to reduce hallucinations and using dynamic weight scheduling to mitigate data imbalance.

A core contribution of this work is a comparative study of encoder architectures, featuring the first application of nGPT to speech processing. Our experiments reveal a crucial trade-off: while the final FastConformer-based model excelled after a multi-stage training and fine-tuning regimen , the nGPT encoder is a data-hungry model that demonstrated stronger and faster generalization in large-scale, single-stage training, especially on challenging translation tasks. This investigation also yielded novel insights into positional encodings, finding ALiBi more effective for long-form ASR and RoPE more suitable for AST.

We also present a robust method for generating accurate segment-level timestamps using the NeMo Forced Aligner , and the release of Parakeet-TDT-0.6B-v3 extends high-quality ASR to resource-limited settings. 

%Bibliography
\bibliographystyle{nemo_speech}  
\bibliography{nemo_speech}  

\clearpage
\appendix

\section*{Appendix}
\addcontentsline{toc}{section}{Appendix}
\nopagebreak[4]            % <-- prevent a break here

% \section{Training Data Overview}
\label{app:training_data_overview}
\begin{table}[ht]
  \setlength{\abovecaptionskip}{10pt} % add space between table and caption
  \label{tab:data-by-corpus-plain}
  \begin{tabular}{lrrrr}
    \hline
    Corpus & ASR & X$\to$En & En$\to$X & Total \\
    \hline
    Granary          & 638,229.73 & 358,284.40 & -       & 996,514.13 \\
    NeMo ASR Set 3.0 & 19,843.10  & 10,165.01  & 197,020.66 & 227,028.77 \\
    Supplementary dataset & -       & -       & 477,739.85 & 477,739.85 \\
    \hline
    \textbf{Total}   & 658,072.83 & 368,449.41 & 674,760.51 & 1,701,282.75 \\
    \multicolumn{4}{l}{\textbf{Non-speech data}} & 36,581.15 \\
    \textbf{Grand Total} & \multicolumn{3}{c}{} & \textbf{1,737,863.91} \\
    \hline
  \end{tabular}
  \centering
  \caption{Training data hours by corpus and tasks involved.}
\end{table}

\section{Training Data Language, Corpora and Task Decomposition}\label{app:lang_dist}
\begin{table}[ht]
\centering
\scriptsize
\setlength{\tabcolsep}{4pt}
\setlength{\abovecaptionskip}{10pt} % add space between table and caption
\renewcommand{\arraystretch}{1.1}
\begin{tabular}{l|cc|cc|cc}
\hline
\multirow{2}{*}{Lang} & \multicolumn{2}{c|}{ASR} & \multicolumn{2}{c|}{X$\to$En} & \multicolumn{2}{c}{En$\to$X} \\
 & Granary & NeMo & Granary & NeMo & NeMo & Supplementary \\
\hline
bg & 13986.55 & 9.49 & 13875.61 & 9.30 & 8346.43 & 20194.40 \\
cs & 15043.85 & 58.56 & 14936.10 & 57.92 & 8408.30 & 20231.86 \\
da & 10104.41 & 13.08 & 10068.36 & 12.80 & 8041.09 & 20113.59 \\
de & 29279.61 & 2602.24 & 28974.96 & 2509.15 & 8781.31 & 20267.06 \\
el & 11008.78 & 24.89 & 10927.85 & 23.20 & 8472.81 & 20259.95 \\
es & 45812.67 & 1505.47 & 45353.29 & 1406.23 & 8505.52 & 20272.88 \\
et & 7983.65 & 10.54 & 7942.66 & 10.44 & 8453.49 & 20129.90 \\
fi & 10856.40 & 17.32 & 10841.47 & 17.00 & 8484.42 & 20262.53 \\
fr & 39226.50 & 1989.65 & 38626.92 & 1998.26 & 8482.86 & 20259.14 \\
hr & 5285.79 & 1671.12 & 5273.91 & 1646.91 & 5119.01 & 14850.72 \\
hu & 11818.71 & 65.15 & 11729.05 & 64.46 & 8452.52 & 20234.48 \\
it & 22962.30 & 515.16 & 22896.12 & 521.30 & 8515.96 & 20269.42 \\
lt & 10775.43 & 20.10 & 10731.04 & 19.70 & 8410.20 & 20213.40 \\
lv & 9311.46 & 8.58 & 9213.82 & 8.46 & 8663.14 & 20145.42 \\
mt & 4009.81 & 13.98 & 3524.11 & 13.67 & 7078.29 & 19519.90 \\
nl & 13997.41 & 9.64 & 13957.39 & 14.79 & 8477.62 & 20275.66 \\
pl & 17202.73 & 316.13 & 17071.60 & 308.58 & 8505.21 & 20272.97 \\
pt & 29869.11 & 16.99 & 29505.53 & 20.22 & 8478.27 & 20284.43 \\
ro & 12419.03 & 21.49 & 12368.69 & 21.20 & 8451.23 & 20253.89 \\
ru & 20460.39 & 1716.46 & 19595.31 & 1263.28 & 8511.02 & 20262.18 \\
sk & 4467.62 & 22.54 & 4439.20 & 21.67 & 7475.55 & 19374.72 \\
sl & 5851.59 & 9.41 & 5826.37 & 9.53 & 7688.19 & 19301.57 \\
sv & 10014.93 & 10.09 & 9991.90 & 9.88 & 8735.76 & 20253.51 \\
uk & 932.67 & 191.06 & 613.14 & 177.06 & 8482.46 & 20236.27 \\
en & 275548.32 & 9003.94 & -- & -- & -- & -- \\
\hline
\end{tabular}
\caption{Training data (hours) by language, corpora, and task. NeMo = NeMo ASR Set 3.0.}
\label{tab:big-multicol}
\end{table}

\begin{landscape}
\section{Evaluation: ASR Performance}\label{app:eval_details_asr}
\begin{table}[ht]
\centering
\setlength{\abovecaptionskip}{10pt} % add space between table and caption
\scalebox{0.65}{%
\begin{tabular}{l*{25}{c}}
\toprule
\textbf{Model} & \textbf{bg} & \textbf{cs} & \textbf{da} & \textbf{de} & \textbf{el} & \textbf{en} & \textbf{es} & \textbf{et} & \textbf{fi} & \textbf{fr} & \textbf{hr} & \textbf{hu} & \textbf{it} & \textbf{lt} & \textbf{lv} & \textbf{mt} & \textbf{nl} & \textbf{pl} & \textbf{pt} & \textbf{ro} & \textbf{ru} & \textbf{sk} & \textbf{sl} & \textbf{sv} & \textbf{uk} \\
\midrule
whisper-large-v3      & 12.86 & 11.33 & 12.57 & \textbf{4.30} & 27.03 & 4.25 & 3.12 & 19.12 & \textbf{7.70} & 6.31 & 11.07 & 14.11 & \textbf{2.31} & 22.34 & 18.29 & 68.89 & \textbf{5.57} & \textbf{4.74} & \textbf{3.65} & 8.24 & \textbf{4.17} & 8.40 & 19.93 & \textbf{7.88} & \textbf{6.51} \\
seamless-m4t-v2-large & 10.67 & 9.31 & 12.09 & 6.25 & 25.68 & 6.15 & 5.10 & \textbf{11.75} & 12.24 & 6.82 & 11.89 & \textbf{11.38} & 4.49 & 15.47 & -- & \textbf{18.16} & 8.15 & 8.76 & 7.32 & 8.87 & 7.43 & 7.42 & \textbf{11.18} & 11.13 & 9.07 \\
seamless-m4t-medium          & 14.42 & 14.89 & 18.26 & 9.77 & 29.49 & 6.99 & 6.79 & 20.99 & 21.09 & 10.49 & 14.16 & 24.09 & 6.72 & 27.42 & -- & 23.03 & 11.96 & 14.84 & 8.87 & 13.95 & 12.82 & 11.16 & 20.38 & 18.13 & 14.55 \\
Voxtral-Mini-3B-2507  & -- & -- & -- & 4.62 & -- & 3.88 & 3.34 & -- & -- & \textbf{4.63} & -- & -- & 2.49 & -- & -- & -- & -- & -- & 3.74 & -- & -- & -- & -- & -- & -- \\
Phi-4-multimodal-instruct    & -- & -- & -- & 4.37 & -- & \textbf{3.64} & 3.23 & -- & -- & 4.64 & -- & -- & 2.48 & -- & -- & -- & -- & -- & 4.00 & -- & -- & -- & -- & -- & -- \\
\midrule
\emph{Parakeet-TDT-0.6B-v3}          & 12.64 & 11.01 & 18.41 & 5.04 & 20.70 & 4.85 & 3.45 & 17.73 & 13.21 & 5.15 & 12.46 & 15.72 & 3.00 & 20.35 & 22.84 & 20.46 & 7.48 & 7.31 & 4.76 & 12.44 & 5.51 & 8.82 & 24.03 & 15.08 & 6.79 \\
\midrule
\emph{Canary-1B-v2}          & \textbf{9.25} & \textbf{7.86} & \textbf{11.25} & 4.40 & \textbf{9.21} & 4.50 & \textbf{2.90} & 12.55 & 8.59 & 5.02 & \textbf{8.29} & 12.90 & 3.07 & \textbf{12.36} & \textbf{9.66} & 18.31 & 6.12 & 6.64 & 4.39 & \textbf{6.61} & 6.90 & \textbf{5.74} & 13.32 & 9.57 & 10.50 \\
\bottomrule
\end{tabular}}
\caption{Per-language WER (\%) results on the \textbf{FLEURS} benchmark.}
\label{tab:fleurs-results}
\end{table}

\begin{table}[ht]
\centering
\setlength{\abovecaptionskip}{10pt} % add space between table and caption
\begin{tabular}{l*{6}{c}}
\toprule
\textbf{Model} & \textbf{es} & \textbf{fr} & \textbf{it} & \textbf{nl} & \textbf{pl} & \textbf{pt} \\
\midrule
whisper-large-v3      & 4.89 & 7.15 & 9.26 & 12.08 & 8.88 & 5.96 \\
seamless-m4t-v2-large & 3.08 & 3.73 & \textbf{6.83} & \textbf{6.91} & \textbf{5.74} & \textbf{5.86} \\
seamless-m4t-medium          & 4.43 & 6.08 & 10.13 & 8.57 & 10.22 & 9.36 \\
Voxtral-Mini-3B-2507  & 3.85 & 5.75 & 9.36 & -- & -- & 5.87 \\
Phi-4-multimodal-instruct    & 3.72 & 4.13 & 8.51 & -- & -- & 6.31 \\
\midrule
\emph{Parakeet-TDT-0.6B-v3}          & 4.39 & 4.97 & 10.08 & 12.78 & 7.28 & 7.50 \\
\midrule
\emph{Canary-1B-v2}          & \textbf{2.94} & \textbf{3.36} & 9.16 & 11.27 & 8.77 & 8.14 \\
\bottomrule
\end{tabular}
\caption{Per-language WER (\%) results on the \textbf{MLS} benchmark.}
\label{tab:mls-results}
\end{table}

\begin{table}[ht]
\centering
\setlength{\abovecaptionskip}{10pt} % add space between table and caption
\scalebox{0.65}{%
\begin{tabular}{l*{13}{c}}
\toprule
\textbf{Model} & \textbf{de} & \textbf{en} & \textbf{es} & \textbf{et} & \textbf{fr} & \textbf{it} & \textbf{lv} & \textbf{nl} & \textbf{pt} & \textbf{ru} & \textbf{sl} & \textbf{sv} & \textbf{uk} \\
\midrule
whisper-large-v3      & 5.94 & 8.52 & 4.32 & 28.54 & 11.02 & 5.45 & 18.84 & \textbf{5.97} & \textbf{3.72} & 4.04 & 16.63 & \textbf{8.06} & 12.52 \\
seamless-m4t-v2-large & 5.34 & \textbf{5.01} & 4.20 & \textbf{9.80} & 6.44 & 4.46 & -- & \textbf{5.97} & 3.87 & \textbf{2.56} & \textbf{5.39} & 9.20 & 9.01 \\
seamless-m4t-medium          & 9.83 & 7.80 & 6.95 & 22.05 & 11.86 & 7.56 & -- & 13.78 & 8.62 & 7.15 & 14.38 & 19.89 & 17.51 \\
Voxtral-Mini-3B-2507  & 6.09 & 7.60 & 4.23 & -- & 7.51 & 5.80 & -- & -- & -- & -- & -- & -- & -- \\
Phi-4-multimodal-instruct    & \textbf{4.62} & 5.83 & 3.81 & -- & 6.61 & \textbf{3.42} & -- & -- & -- & -- & -- & -- & -- \\
\midrule
\emph{Parakeet-TDT-0.6B-v3}          & 4.84 & 6.80 & \textbf{3.41} & 22.04 & \textbf{6.05} & 3.69 & 38.36 & 6.50 & 3.96 & 3.00 & 31.80 & 20.16 & \textbf{5.10} \\
\midrule
\emph{Canary-1B-v2}          & 5.53 & 6.85 & 3.81 & 18.28 & 6.30 & 4.80 & \textbf{11.49} & 6.93 & 6.87 & 5.14 & 7.59 & 13.32 & 18.15 \\
\bottomrule
\end{tabular}}
\caption{Per-language WER (\%) results on the \textbf{CoVoST2} benchmark.}
\label{tab:covost2-results}
\end{table}

\clearpage
\section{\texorpdfstring{Evaluation: AST (X$\to$En) Performance}{Evaluation: AST (X→En) Performance}}
\label{app:eval_details_ast_x-en}

\begin{table}[ht]
\centering
\setlength{\abovecaptionskip}{10pt}
\scalebox{0.65}{%
\begin{tabular}{l*{24}{cc}}
\toprule
& \multicolumn{2}{c}{\textbf{bg}} & \multicolumn{2}{c}{\textbf{cs}} & \multicolumn{2}{c}{\textbf{da}} & \multicolumn{2}{c}{\textbf{de}} & \multicolumn{2}{c}{\textbf{el}} & \multicolumn{2}{c}{\textbf{es}} & \multicolumn{2}{c}{\textbf{et}} & \multicolumn{2}{c}{\textbf{fi}} & \multicolumn{2}{c}{\textbf{fr}} & \multicolumn{2}{c}{\textbf{hr}} & \multicolumn{2}{c}{\textbf{hu}} & \multicolumn{2}{c}{\textbf{it}} \\
\cmidrule(lr){2-3} \cmidrule(lr){4-5} \cmidrule(lr){6-7} \cmidrule(lr){8-9} \cmidrule(lr){10-11} \cmidrule(lr){12-13} \cmidrule(lr){14-15} \cmidrule(lr){16-17} \cmidrule(lr){18-19} \cmidrule(lr){20-21} \cmidrule(lr){22-23} \cmidrule(lr){24-25}
\textbf{Model} & C & B & C & B & C & B & C & B & C & B & C & B & C & B & C & B & C & B & C & B & C & B & C & B \\
\midrule
whisper-large-v3 & 77.61 & 26.93 & 76.62 & 25.56 & 78.68 & 31.50 & 81.68 & 32.35 & 74.56 & 21.24 & 80.81 & 21.86 & 69.88 & 17.30 & 77.59 & 19.86 & 81.83 & 31.00 & 77.42 & 25.95 & 72.98 & 18.33 & 81.46 & 22.65 \\
seamless-m4t-v2-large      & \textbf{80.48} & \textbf{33.03} & \textbf{82.56} & \textbf{34.36} & \textbf{82.99} & \textbf{37.90} & 83.98 & 37.38 & \textbf{80.20} & \textbf{27.99} & 81.94 & 25.56 & \textbf{83.52} & \textbf{32.00} & \textbf{84.51} & \textbf{28.57} & 82.77 & 33.77 & \textbf{81.86} & \textbf{31.21} & \textbf{81.99} & \textbf{28.13} & 82.92 & 26.76 \\
seamless-m4t-medium  & 78.23 & 27.96 & 77.91 & 27.16 & 78.62 & 31.95 & 81.57 & 33.67 & 76.32 & 22.80 & 79.10 & 21.92 & 78.39 & 24.07 & 79.11 & 22.12 & 80.35 & 30.66 & 78.70 & 26.84 & 72.52 & 18.51 & 80.08 & 23.43 \\
Voxtral-Mini-3B-2507 & --    & --    & --    & --    & --    & --    & \textbf{85.01} & \textbf{38.47} & --    & --    & \textbf{83.73} & \textbf{28.32} & --    & --    & --    & --    & \textbf{84.33} & \textbf{35.84} & --    & --    & --    & --    & \textbf{84.19} & \textbf{29.31} \\
Phi-4-multimodal-instruct      & --    & --    & --    & --    & --    & --    & 84.93 & 38.16 & --    & --    & 82.85 & 25.66 & --    & --    & --    & --    & 83.88 & 35.28 & --    & --    & --    & --    & 83.62 & 26.44 \\
\midrule
\emph{Canary-1B-v2}          & 79.60 & 30.93 & 78.64 & 29.28 & 80.45 & 34.80 & 83.09 & 36.03 & 76.73 & 24.08 & 81.19 & 25.45 & 80.25 & 28.38 & 80.81 & 24.68 & 82.80 & 34.10 & 78.48 & 29.09 & 76.86 & 24.26 & 82.03 & 25.57 \\
\bottomrule
\end{tabular}
}

\vspace{0.6em}

\scalebox{0.65}{%
\begin{tabular}{l*{24}{cc}}
\toprule
& \multicolumn{2}{c}{\textbf{lt}} & \multicolumn{2}{c}{\textbf{lv}} & \multicolumn{2}{c}{\textbf{mt}} & \multicolumn{2}{c}{\textbf{nl}} & \multicolumn{2}{c}{\textbf{pl}} & \multicolumn{2}{c}{\textbf{pt}} & \multicolumn{2}{c}{\textbf{ro}} & \multicolumn{2}{c}{\textbf{ru}} & \multicolumn{2}{c}{\textbf{sk}} & \multicolumn{2}{c}{\textbf{sl}} & \multicolumn{2}{c}{\textbf{sv}} & \multicolumn{2}{c}{\textbf{uk}} \\
\cmidrule(lr){2-3} \cmidrule(lr){4-5} \cmidrule(lr){6-7} \cmidrule(lr){8-9} \cmidrule(lr){10-11} \cmidrule(lr){12-13} \cmidrule(lr){14-15} \cmidrule(lr){16-17} \cmidrule(lr){18-19} \cmidrule(lr){20-21} \cmidrule(lr){22-23} \cmidrule(lr){24-25}
\textbf{Model} & C & B & C & B & C & B & C & B & C & B & C & B & C & B & C & B & C & B & C & B & C & B & C & B \\
\midrule
whisper-large-v3 & 65.05 & 13.69 & 66.11 & 15.40 & 54.96 & 12.18 & 79.57 & 22.12 & 77.43 & 20.90 & 83.22 & 37.63 & 81.07 & 30.34 & 80.72 & 26.56 & 76.70 & 23.35 & 70.12 & 17.27 & 81.07 & 34.28 & 78.71 & 27.95 \\
seamless-m4t-v2-large      & \textbf{77.98} & \textbf{24.53} & -- & -- & \textbf{74.02} & \textbf{41.18} & \textbf{82.23} & \textbf{28.01} & \textbf{79.98} & \textbf{24.85} & 82.62 & 39.39 & \textbf{83.21} & \textbf{35.26} & \textbf{81.85} & \textbf{30.27} & \textbf{82.13} & \textbf{33.03} & \textbf{80.52} & \textbf{26.62} & \textbf{83.13} & \textbf{38.05} & \textbf{81.86} & \textbf{33.34} \\
seamless-m4t-medium  & 69.29 & 17.00 & --    & --    & 70.43 & 34.68 & 79.00 & 23.07 & 74.33 & 18.73 & 80.18 & 34.72 & 78.83 & 29.58 & 78.29 & 24.03 & 78.94 & 27.07 & 72.79 & 19.45 & 77.61 & 30.71 & 77.20 & 25.76 \\
Voxtral-Mini-3B-2507 & --    & --    & --    & --    & --    & --    & --    & --  & --    & --  & \textbf{85.37} & 43.00     & --    & --    & --    & --    & --    & --    & --    & --    & --    & --    & --    & -- \\
Phi-4-multimodal-instruct      & --    & --    & --    & --    & --    & --    & --    & --  & --    & --   & 85.01 & 41.41    & --    & --    & --    & --    & --    & --    & --    & --    & --    & --    & --    & -- \\
\midrule
\emph{Canary-1B-v2}          & 76.30 & 22.86 & \textbf{79.71} & \textbf{27.86} & 70.00 & 34.99 & 80.72 & 26.49 & 77.05 & 22.30 & 82.91 & \textbf{39.43} & 81.61 & 33.55 & 79.17 & 27.26 & 79.86 & 30.55 & 76.89 & 23.65 & 80.75 & 34.92 & 77.23 & 27.50 \\
\bottomrule
\end{tabular}
}
\caption{Per-language COMET (C) and BLEU (B) scores (\%) on the \textbf{FLEURS} benchmark.}
\label{tab:fleurs-comet-bleu_x-en}
\end{table}

% Preamble: \usepackage{booktabs,graphicx}
\begin{table}[ht]
\centering
\setlength{\abovecaptionskip}{10pt}
\scalebox{0.65}{%
\begin{tabular}{l*{11}{cc}}
\toprule
& \multicolumn{2}{c}{\textbf{de}} & \multicolumn{2}{c}{\textbf{es}} & \multicolumn{2}{c}{\textbf{et}} & \multicolumn{2}{c}{\textbf{fr}} & \multicolumn{2}{c}{\textbf{it}} & \multicolumn{2}{c}{\textbf{lv}} & \multicolumn{2}{c}{\textbf{nl}} & \multicolumn{2}{c}{\textbf{pt}} & \multicolumn{2}{c}{\textbf{ru}} & \multicolumn{2}{c}{\textbf{sl}} & \multicolumn{2}{c}{\textbf{sv}} \\
\cmidrule(lr){2-3} \cmidrule(lr){4-5} \cmidrule(lr){6-7} \cmidrule(lr){8-9} \cmidrule(lr){10-11} \cmidrule(lr){12-13} \cmidrule(lr){14-15} \cmidrule(lr){16-17} \cmidrule(lr){18-19} \cmidrule(lr){20-21} \cmidrule(lr){22-23}
\textbf{Model} & C & B & C & B & C & B & C & B & C & B & C & B & C & B & C & B & C & B & C & B & C & B \\
\midrule
whisper-large-v3         & 75.28 & 34.57 & 78.52 & 39.82 & 63.98 & 13.93 & 74.38 & 35.73 & 76.55 & 36.35 & 58.63 & 16.93 & 77.46 & 40.53 & 79.58 & 51.14 & 80.99 & 41.31 & 64.61 & 24.34 & 74.70 & 44.19 \\
seamless-m4t-v2-large              & \textbf{79.62} & \textbf{40.40} & \textbf{81.29} & 43.47 & \textbf{80.25} & \textbf{27.52} & \textbf{78.95} & \textbf{42.36} & 80.05 & 40.49 & --    & --    & \textbf{80.77} & \textbf{43.90} & 80.89 & \textbf{55.18} & \textbf{84.74} & \textbf{53.78} & \textbf{78.19} & \textbf{42.44} & \textbf{77.87} & \textbf{46.78} \\
seamless-m4t-medium      & 75.74 & 35.76 & 78.65 & 39.63 & 73.72 & 22.84 & 76.47 & 39.23 & 77.74 & 37.64 & --    & --    & 75.24 & 37.35 & 76.62 & 47.37 & 81.06 & 43.20 & 70.50 & 32.15 & 70.70 & 36.59 \\
Voxtral-Mini-3B-2507     & 77.83 & 36.35 & 80.53 & 41.84 & --    & --    & 77.39 & 38.93 & 78.79 & 38.36 & --    & --    & --    & --    & 79.95 & 51.03 & --    & --    & --    & --    & --    & --    \\
Phi-4-multimodal-instruct & 79.46 & 40.01 & 81.10 & \textbf{43.53} & --    & --    & 78.57 & 41.98 & \textbf{80.20} & \textbf{41.29} & --    & --    & --    & --    & \textbf{81.20} & 54.52 & --    & --    & --    & --    & --    & --    \\
\midrule
\emph{Canary-1B-v2}                  & 78.32 & 39.22 & 80.82 & 42.74 & 75.78 & 25.52 & 78.52 & 41.43 & 79.45 & 40.03 & \textbf{70.91} & \textbf{31.77} & 78.46 & 41.59 & 78.26 & 50.38 & 83.31 & 48.78 & 74.72 & 39.43 & 73.71 & 44.40 \\
\bottomrule
\end{tabular}
}
\caption{Per-language COMET (C) and BLEU (B) scores (\%) on the \textbf{CoVoST2} benchmark.}
\label{tab:covost2-comet-bleu-full}
\end{table}

\clearpage
\section{\texorpdfstring{Evaluation: AST (En$\to$X) Performance}{Evaluation: AST (En→X) Performance}}
\label{app:eval_details_ast_en-x}

\begin{table}[ht]
\centering
\setlength{\abovecaptionskip}{10pt}
\scalebox{0.65}{%
\begin{tabular}{l*{24}{cc}}
\toprule
& \multicolumn{2}{c}{\textbf{bg}} & \multicolumn{2}{c}{\textbf{cs}} & \multicolumn{2}{c}{\textbf{da}} & \multicolumn{2}{c}{\textbf{de}} & \multicolumn{2}{c}{\textbf{el}} & \multicolumn{2}{c}{\textbf{es}} & \multicolumn{2}{c}{\textbf{et}} & \multicolumn{2}{c}{\textbf{fi}} & \multicolumn{2}{c}{\textbf{fr}} & \multicolumn{2}{c}{\textbf{hr}} & \multicolumn{2}{c}{\textbf{hu}} & \multicolumn{2}{c}{\textbf{it}} \\
\cmidrule(lr){2-3} \cmidrule(lr){4-5} \cmidrule(lr){6-7} \cmidrule(lr){8-9} \cmidrule(lr){10-11} \cmidrule(lr){12-13} \cmidrule(lr){14-15} \cmidrule(lr){16-17} \cmidrule(lr){18-19} \cmidrule(lr){20-21} \cmidrule(lr){22-23} \cmidrule(lr){24-25}
\textbf{Model} & C & B & C & B & C & B & C & B & C & B & C & B & C & B & C & B & C & B & C & B & C & B & C & B \\
\midrule
seamless-m4t-v2-large      & 87.53 & 37.01 & \textbf{86.83} & 27.36 & \textbf{87.06} & 41.07 & \textbf{83.86} & 34.40 & \textbf{84.80} & 16.70 & 81.66 & 24.40 & 87.02 & 23.38 & \textbf{88.18} & 21.03 & 83.59 & \textbf{44.81} & \textbf{86.67} & \textbf{25.48} & \textbf{85.51} & \textbf{21.60} & 83.87 & 25.28 \\
seamless-m4t-medium  & 83.32 & 30.63 & 80.70 & 21.65 & 82.62 & 34.93 & 77.69 & 27.94 & 80.85 & 13.82 & 78.50 & 21.30 & 80.11 & 16.19 & 80.63 & 14.18 & 79.07 & 37.98 & 81.36 & 20.55 & 78.37 & 15.69 & 79.75 & 21.90 \\
Voxtral-Mini-3B-2507 & --    & --    & --    & --    & --    & --    & 83.25 & 31.24 & --    & --    & \textbf{82.37} & 24.45 & --    & --    & --    & --    & 83.66 & 41.69 & --    & --    & --    & --    & 83.92 & 24.83 \\
Phi-4-multimodal-instruct      & --    & --    & --    & --    & --    & --    & 83.57 & \textbf{35.57} & --    & --    & 80.90 & 24.27 & --    & --    & --    & --    & 81.88 & 39.61 & --    & --    & --    & --    & 81.99 & 24.39 \\
\midrule
\emph{Canary-1B-v2}          & \textbf{87.73} & \textbf{38.14} & 86.26 & \textbf{27.69} & 86.89 & \textbf{41.78} & 83.30 & 33.65 & 81.49 & \textbf{23.87} & 82.13 & \textbf{25.67} & \textbf{87.32} & \textbf{23.54} & 87.40 & \textbf{21.10} & \textbf{83.82} & 43.42 & 85.46 & 24.71 & 83.94 & 20.75 & \textbf{84.12} & \textbf{26.82} \\
\bottomrule
\end{tabular}
}

\vspace{0.6em}

\scalebox{0.65}{%
\begin{tabular}{l*{24}{cc}}
\toprule
& \multicolumn{2}{c}{\textbf{lt}} & \multicolumn{2}{c}{\textbf{lv}} & \multicolumn{2}{c}{\textbf{mt}} & \multicolumn{2}{c}{\textbf{nl}} & \multicolumn{2}{c}{\textbf{pl}} & \multicolumn{2}{c}{\textbf{pt}} & \multicolumn{2}{c}{\textbf{ro}} & \multicolumn{2}{c}{\textbf{ru}} & \multicolumn{2}{c}{\textbf{sk}} & \multicolumn{2}{c}{\textbf{sl}} & \multicolumn{2}{c}{\textbf{sv}} & \multicolumn{2}{c}{\textbf{uk}} \\
\cmidrule(lr){2-3} \cmidrule(lr){4-5} \cmidrule(lr){6-7} \cmidrule(lr){8-9} \cmidrule(lr){10-11} \cmidrule(lr){12-13} \cmidrule(lr){14-15} \cmidrule(lr){16-17} \cmidrule(lr){18-19} \cmidrule(lr){20-21} \cmidrule(lr){22-23} \cmidrule(lr){24-25}
\textbf{Model} & C & B & C & B & C & B & C & B & C & B & C & B & C & B & C & B & C & B & C & B & C & B & C & B \\
\midrule
seamless-m4t-v2-large      & \textbf{85.42} & 20.49 & -- & -- & 68.89 & \textbf{32.45} & 84.14 & 24.61 & \textbf{84.52} & \textbf{18.88} & 84.83 & 43.83 & 86.41 & 35.05 & \textbf{85.18} & \textbf{27.55} & \textbf{86.40} & \textbf{29.04} & \textbf{85.91} & 24.57 & \textbf{87.10} & \textbf{41.53} & \textbf{86.21} & 25.55 \\
seamless-m4t-medium  & 78.96 & 15.85 & --    & --    & 67.57 & 27.26 & 80.02 & 21.44 & 78.72 & 14.27 & 81.97 & 37.91 & 81.61 & 28.76 & 80.60 & 22.24 & 79.73 & 22.24 & 78.63 & 18.65 & 82.39 & 34.45 & 80.66 & 20.82 \\
Voxtral-Mini-3B-2507 & --    & --    & --    & --    & --    & --    & --    & --    & --    & -- & 84.66 & 39.74     & --    & --    & --    & --    & --    & --    & --    & --    & --    & --    & -- & -- \\
Phi-4-multimodal-instruct      & --    & --    & --    & --    & --    & --    & --    & --  & --    & --   & 82.65 & 38.43     & --    & --    & --    & --    & --    & --    & --    & --    & --    & --    & -- & -- \\
\midrule
\emph{Canary-1B-v2}          & 85.13 & \textbf{21.60} & \textbf{86.52} & \textbf{29.33} & \textbf{69.02} & 31.61 & \textbf{84.25} & \textbf{25.81} & 83.82 & 17.98 & \textbf{85.56} & \textbf{44.75} & \textbf{87.00} & \textbf{36.27} & 84.87 & 27.21 & 86.21 & 28.43 & 84.96 & \textbf{24.96} & 86.43 & 40.73 & 85.74 & \textbf{25.72} \\
\bottomrule
\end{tabular}
}
\caption{Per-language COMET (C) and BLEU (B) scores (\%) on the \textbf{FLEURS} benchmark (new evaluation).}
\label{tab:fleurs-comet-bleu}
\end{table}

\begin{table}[ht]
\centering
\setlength{\abovecaptionskip}{10pt}
\scalebox{0.7}{%
\begin{tabular}{l*{5}{cc}}
\toprule
& \multicolumn{2}{c}{\textbf{de}} & \multicolumn{2}{c}{\textbf{et}} & \multicolumn{2}{c}{\textbf{lv}} & \multicolumn{2}{c}{\textbf{sl}} & \multicolumn{2}{c}{\textbf{sv}} \\
\cmidrule(lr){2-3} \cmidrule(lr){4-5} \cmidrule(lr){6-7} \cmidrule(lr){8-9} \cmidrule(lr){10-11}
\textbf{Model} & C & B & C & B & C & B & C & B & C & B \\
\midrule
seamless-m4t-v2-large      & \textbf{81.30} & \textbf{37.36} & \textbf{82.89} & \textbf{29.57} & -- & -- & \textbf{83.10} & \textbf{36.53} & \textbf{83.34} & \textbf{43.71} \\
seamless-m4t-medium  & 77.20 & 32.93 & 78.52 & 24.76 & -- & -- & 78.80 & 30.62 & 79.86 & 39.46 \\
Voxtral-Mini-3B-2507 & 76.51 & 28.78 & -- & -- & -- & -- & -- & -- & -- & -- \\
Phi-4-multimodal-instruct      & 79.33 & 34.12 & -- & -- & -- & -- & -- & -- & -- & -- \\
\midrule
\emph{Canary-1B-v2}          & 78.37 & 33.82 & 80.61 & 28.09 & \textbf{81.32} & \textbf{27.10} & 80.02 & 31.18 & 81.12 & 41.49 \\
\bottomrule
\end{tabular}
}
\caption{Per-language COMET (C) and BLEU (B) scores (\%) on the \textbf{CoVoST2} benchmark.}
\label{tab:covost2-comet-bleu}
\end{table}

\end{landscape}

 % no forced page break

\end{document}